\pdfoutput=1
\documentclass{article}
\usepackage{iclr2027_conference,times}
\iclrfinalcopy  %

\usepackage{amsmath,amssymb,booktabs,graphicx,xcolor,multirow}
\usepackage{hyperref}
\usepackage{url}
\usepackage{tabularx}
\usepackage{caption}
\usepackage{makecell}
\usepackage{tikz}
\usepackage{pgfplots}
\pgfplotsset{compat=1.18}
\usetikzlibrary{arrows.meta,positioning,calc,fit,backgrounds,
  decorations.pathreplacing,shapes.geometric,shapes.misc}
\usepgfplotslibrary{groupplots}

\definecolor{cTeach}{HTML}{2A78D6}
\definecolor{cData}{HTML}{EB6834}
\definecolor{cArith}{HTML}{1BAF7A}
\definecolor{cInk}{HTML}{1F2933}
\definecolor{cMuted}{HTML}{6B7280}
\definecolor{cGrid}{HTML}{D9DCE1}
\definecolor{cFrozen}{HTML}{E9EDF1}

\tikzset{
  >={Stealth[length=4pt,width=3.2pt]},
  fig/.style={font=\scriptsize, line join=round, line cap=round, cInk},
  box/.style={draw=cInk, rounded corners=2pt, line width=0.5pt, align=center,
    inner sep=2.5pt, minimum height=0.55cm},
  frozen/.style={box, fill=cFrozen},
  teach/.style={box, draw=cTeach, fill=cTeach!10},
  data/.style={box, draw=cData, fill=cData!10},
  arith/.style={box, draw=cArith!80!black, fill=cArith!10},
  flow/.style={->, line width=0.6pt, cInk},
  sup/.style={->, line width=0.9pt, dashed},
  panel/.style={font=\scriptsize\bfseries, anchor=north west, inner sep=0pt},
}

\pgfplotsset{
  paperplot/.style={
    font=\scriptsize,
    axis line style={cMuted, line width=0.4pt},
    tick style={cMuted, line width=0.4pt},
    tick label style={font=\tiny, cInk},
    label style={font=\scriptsize, cInk},
    title style={font=\scriptsize\bfseries, cInk, yshift=-3pt},
    major grid style={cGrid, line width=0.3pt},
    ymajorgrids,
    axis x line*=bottom, axis y line*=left,
    legend style={font=\tiny, draw=none, fill=none, cells={anchor=west}},
    legend cell align=left,
    every axis plot/.append style={line width=0.9pt},
  },
}

\usepackage{mathtools}
\hypersetup{
  pdftitle={Action Jets: Learning the Local Shape of the Action-conditioned Data Manifold},
  pdfauthor={Anton Arapin, Chahyon Ku, Bernadette Bucher}}
\DeclareRobustCommand{\pending}[1]{%
  \PackageWarning{main_arxiv}{\string\pending\space marker left in the text}%
  \textcolor{red}{[#1]}}
\newcommand{\J}{\widehat{J}}
\newcommand{\R}{\mathbb{R}}
\newcommand{\E}{\mathbb{E}}
\newcommand{\norm}[1]{\lVert #1 \rVert}

\title{Action Jets: Learning the Local Shape of the Action-conditioned Data Manifold}

\author{Anton Arapin,\thanks{\begin{tabular}[t]{@{}l@{}}Correspondence to: Anton Arapin \textless\href{mailto:aarapin@umich.edu}{aarapin@umich.edu}\textgreater\\ Project page: \url{https://anton-arapin.github.io/action-jets-site}\end{tabular}}\hspace{0.6em}Chahyon Ku, Bernadette Bucher \\
University of Michigan
}

\begin{document}
\maketitle
\lhead{Preprint}

\begin{abstract}
World models are a powerful means of reasoning and planning, but many of these action-conditioned predictive models are high dimensional and computationally expensive. These challenges limit the number of action candidates that can be evaluated by a planner. We show that this cost is avoidable for planners over short horizons, which only require local information with respect to the current state. For a fixed state, the frames reachable by varying a low-dimensional action (a camera movement or an end-effector command) lie on a low-dimensional manifold. We estimate the local shape of that manifold directly via a network trained to estimate how each image region moves per unit of each action axis supervised by data collected from samples by the generator. A correction head fit to trajectory data estimates the shape of this manifold region up to a third-order Taylor polynomial. We call this network estimating the Taylor polynomial describing the manifold shape an \emph{Action Jet} and compute its radius of validity in closed form. By traversing this approximation of the local manifold shape, we only have to perform inference once per state for any number of candidate actions.
Scoring sixteen candidates from a single extraction is over $300\times$ faster than with either source generator, and this gap widens with the number of candidates a planner evaluates.
We measure against state-of-the-art action-conditioned generators, IRASim and WorldGym on BridgeData V2, a real 7-DoF manipulation benchmark, and FrameCrafter on DL3DV scenes, where we chart from the novel-view generators DepthSplat and LagerNVS.
Because only the tangent comes from the generator and every other term from the data, the Action Jet also predicts held-out frames more accurately than IRASim, the generator that supplied its tangent, at short horizons and at over $100\times$ less compute.
\end{abstract}

\section{Introduction}
\label{sec:intro}

Action-conditioned generative models can act as powerful world models for planning and control, to score candidate actions, to evaluate policies offline, and to plan \citep{zhu2025irasim,quevedo2026worldgym,yang2024learning,bar2025navigation}. However, a planner comparing a hundred candidates performs a hundred model rollouts which are particularly expensive with high dimensional methods like video generators. IRASim~\citep{zhu2025irasim} spends 25 diffusion steps on each candidate action chunk and WorldGym~\citep{quevedo2026worldgym} 40, so this scale is a significant barrier to real-time decision-making.

\begin{figure}[t]
\centering
\def\hs(#1,#2){0.25*(#1)+0.08*(#2)+0.32*(#1)*(#1)-0.22*(#2)*(#2)+0.10*(#1)*(#2)}
\def\jetsurface#1{%
  \foreach \v in {-1.2,-0.6,0,0.6,1.2}{
    \draw[#1, line width=0.42pt] plot[domain=-1.2:1.2, samples=22, smooth]
      (\x,\v,{0.25*\x+0.08*\v+0.32*\x*\x-0.22*\v*\v+0.10*\x*\v});}
  \foreach \u in {-1.2,-0.6,0,0.6,1.2}{
    \draw[#1, line width=0.42pt] plot[domain=-1.2:1.2, samples=22, smooth]
      (\u,\x,{0.25*\u+0.08*\x+0.32*\u*\u-0.22*\x*\x+0.10*\u*\x});}}
\tikzset{proj/.style={x={(1.24cm,-0.27cm)}, y={(0.76cm,0.45cm)}, z={(0cm,1.14cm)}}}
\newcommand{\framelet}[3]{%
  \node[draw=cInk!55, fill=white, line width=0.4pt, minimum width=0.82cm,
        minimum height=0.58cm, inner sep=0] (#1) at #2 {};
  \fill[cGrid!70] ([yshift=-0.10cm]#1.west) rectangle (#1.south east);
  \draw[cInk!40, line width=0.3pt] ([yshift=-0.10cm]#1.west) -- ([yshift=-0.10cm]#1.east);
  \fill[cInk!60] ($(#1.center)+(#3,-0.13)$) rectangle ++(0.13,0.20);
  \draw[cInk!35, line width=0.3pt] ($(#1.north west)+(0.06,-0.07)$) circle (0.045);}
\begin{tikzpicture}[fig]
\node[panel] at (-0.25,1.95) {(a) the manifold $\mathcal{M}_x$};
\begin{scope}[shift={(2.35,-0.15)}, proj]
  \jetsurface{cInk!70}
  \foreach \u/\v in {0.4/0.5, -0.5/0.35, 0.6/-0.4, -0.35/-0.6, 0.9/0.55, -0.85/-0.35, -0.6/0.85, 0.25/-0.9, 0.2/0.3, -0.25/0.65}{
    \filldraw[fill=cData!90!black, draw=white, line width=0.35pt] (\u,\v,{\hs(\u,\v)}) circle (1.4pt);}
  \filldraw[fill=cInk, draw=white, line width=0.35pt] (0,0,0) circle (1.7pt);
  \node[cInk, anchor=north east, inner sep=1pt] at (0,-0.05,-0.02) {\tiny$z_t$};
  \filldraw[fill=cInk!70, draw=white, line width=0.3pt] (-0.85,-0.45,{\hs(-0.85,-0.45)}) circle (1.5pt);
  \filldraw[fill=cInk!70, draw=white, line width=0.3pt] (0.95,-0.5,{\hs(0.95,-0.5)}) circle (1.5pt);
  \coordinate (ptA) at (-0.85,-0.45,{\hs(-0.85,-0.45)});
  \coordinate (ptB) at (0.95,-0.5,{\hs(0.95,-0.5)});
  \coordinate (ptC) at (0,0,0);
  \node[cInk, anchor=west, inner sep=2pt] at (1.2,1.2,{\hs(1.2,1.2)}) {$\mathcal{M}_x$};
\end{scope}
\framelet{frA}{(0.85,-1.55)}{-0.22}
\framelet{frC}{(2.35,-1.55)}{-0.03}
\framelet{frB}{(3.85,-1.55)}{0.16}
\draw[cInk!55, line width=0.35pt] (frA.north) -- (ptA);
\draw[cInk!55, line width=0.35pt] (frC.north) -- (ptC);
\draw[cInk!55, line width=0.35pt] (frB.north east) -- (ptB);
\node[cInk!70, anchor=north, inner sep=2pt, font=\tiny] at (2.35,-1.92) {each point is a latent; $z_t$ is the current frame's};
\node[panel] at (5.35,1.95) {(b) the chart drawn on it};
\begin{scope}[shift={(7.65,-0.15)}, proj]
  \filldraw[fill=cTeach!13, draw=cTeach!55, line width=0.4pt]
    (-1.0,-1.0,{0.25*(-1.0)+0.08*(-1.0)}) -- (1.0,-1.0,{0.25*(1.0)+0.08*(-1.0)}) --
    (1.0,1.0,{0.25*(1.0)+0.08*(1.0)}) -- (-1.0,1.0,{0.25*(-1.0)+0.08*(1.0)}) -- cycle;
  \jetsurface{cInk!45}
  \draw[cInk!60, dashed, line width=0.9pt] plot[domain=-1.2:1.15, samples=28, smooth] (\x,0,{0.25*\x+0.67*\x*\x});
  \draw[white, line width=1.9pt] plot[domain=-1.2:1.15, samples=28, smooth] (\x,0,{0.25*\x+0.32*\x*\x-0.18*\x*\x*\x});
  \draw[cData, line width=1.3pt] plot[domain=-1.2:1.15, samples=28, smooth] (\x,0,{0.25*\x+0.32*\x*\x-0.18*\x*\x*\x});
  \draw[white, line width=2.0pt] (0,-0.55,{-0.08*0.55-0.40*0.3025}) -- (0,0.55,{0.08*0.55-0.40*0.3025});
  \draw[cTeach, line width=1.2pt] (0,-0.55,{-0.08*0.55-0.40*0.3025}) -- (0,0.55,{0.08*0.55-0.40*0.3025});
  \filldraw[fill=cTeach, draw=white, line width=0.35pt] (0,0.55,{0.08*0.55-0.40*0.3025}) circle (1.8pt);
  \filldraw[fill=cTeach, draw=white, line width=0.35pt] (0,-0.55,{-0.08*0.55-0.40*0.3025}) circle (1.8pt);
  \draw[cArith!85!black, line width=1.0pt, dashed] plot[domain=0:360, samples=72, smooth]
    ({0.7*cos(\x)},{0.7*sin(\x)},{\hs(0.7*cos(\x),0.7*sin(\x))});
  \filldraw[fill=cInk, draw=white, line width=0.35pt] (0,0,0) circle (1.6pt);
  \coordinate (zt) at (0,0,0);
  \node[cInk, anchor=north east, inner sep=1pt] at (0,-0.05,-0.02) {\tiny$z_t$};
  \coordinate (ringNW) at ({0.7*cos(125)},{0.7*sin(125)},{\hs(0.7*cos(125),0.7*sin(125))});
  \coordinate (planeF) at (-0.55,-0.95,{0.25*(-0.55)+0.08*(-0.95)});
  \coordinate (genEnd) at (0.95,0,{0.25*0.95+0.67*0.9025});
  \coordinate (jetEnd) at (1.1,0,{0.25*1.1+0.32*1.21-0.18*1.331});
\end{scope}
\node[cInk!70, anchor=south east, inner sep=1pt, font=\tiny] (gl) at (7.85,1.45) {response of $T$};
\draw[cInk!70, line width=0.4pt] (gl.south east) -- (genEnd);
\node[cData!90!black, anchor=north west, inner sep=1pt, font=\tiny] (jl) at (8.95,-0.05) {Action Jet};
\draw[cData!90!black, line width=0.4pt] (jl.north west) -- (jetEnd);
\node[cTeach, anchor=north, inner sep=1pt, font=\tiny] (pl) at (6.95,-1.25) {tangent plane};
\draw[cTeach, line width=0.4pt] (pl.north) -- (planeF);
\node[cArith!70!black, anchor=south east, inner sep=1pt, font=\tiny] (rl) at (6.35,1.3) {radius $r_1^{*}$};
\draw[cArith!70!black, line width=0.4pt] (rl.south east) -- (ringNW);
\node[draw=cInk!55, fill=white, line width=0.4pt, minimum width=0.84cm,
      minimum height=0.60cm, inner sep=0] (tokfr) at (9.55,-1.15) {};
\fill[cGrid!70] ([yshift=-0.10cm]tokfr.west) rectangle (tokfr.south east);
\fill[cInk!60] ($(tokfr.center)+(-0.02,-0.13)$) rectangle ++(0.13,0.20);
\fill[cTeach!45] ($(tokfr.center)+(-0.14,-0.10)$) rectangle ++(0.28,0.20);
\foreach \k in {-1,0,1}{
  \draw[cInk!45, line width=0.3pt] ($(tokfr.north west)+(0.28*\k+0.28,0)$) -- ($(tokfr.south west)+(0.28*\k+0.28,0)$);
  \draw[cInk!45, line width=0.3pt] ($(tokfr.north west)+(0,-0.20*\k-0.20)$) -- ($(tokfr.north east)+(0,-0.20*\k-0.20)$);}
\draw[cTeach!85!black, line width=0.5pt] ($(tokfr.center)+(-0.14,-0.10)$) rectangle ++(0.28,0.20);
\coordinate (tokNE) at ($(tokfr.center)+(0.14,0.10)$);
\coordinate (tokSE) at ($(tokfr.center)+(0.14,-0.10)$);
\draw[white, line width=1.4pt] ($(zt)!0.12!(tokfr.north west)$) -- (tokfr.north west);
\draw[cInk!75, line width=0.45pt] ($(zt)!0.12!(tokfr.north west)$) -- (tokfr.north west);
\node[cInk!70, anchor=north, align=center, inner sep=2pt, font=\tiny] at (9.55,-1.5)
  {the latent $z_t$,\\[-1pt]split into tokens};
\node[panel] at (10.55,1.95) {(c) the chart at that token};
\begin{scope}[shift={(10.8,-0.1)}]
\def\u{0.073}
\coordinate (cTL) at (-0.2,0.95);
\coordinate (cBL) at (-0.2,0.2);
\draw[fill=cTeach!10, draw=cTeach] (0,0.22) rectangle ++(6*\u,16*\u);
\node[font=\tiny] at (3*\u,0.22+16*\u+0.14) {$F$};
\node[font=\tiny, anchor=north, align=center] at (3*\u,0.19) {fields\\[-2pt]$d\!\times\!A$};
\node at (6*\u+0.2,0.22+8*\u) {$=$};
\begin{scope}[shift={(6*\u+0.4,0.22)}]
  \foreach \k in {0,...,11}{\draw[fill=cTeach!18, draw=cTeach!70, line width=0.3pt] (\k*\u,0) rectangle ++(\u,16*\u);}
  \node[font=\tiny] at (6*\u,16*\u+0.14) {$Q$};
  \node[font=\tiny, anchor=north, align=center] at (6*\u,-0.03) {frame\\[-2pt]$d\!\times\!m$};
\end{scope}
\begin{scope}[shift={(6*\u+0.4+12*\u+0.28,0.22)}]
  \draw[fill=cTeach!10, draw=cTeach] (0,0) rectangle ++(6*\u,12*\u);
  \node[font=\tiny] at (3*\u,12*\u+0.14) {$G$};
  \node[font=\tiny, anchor=north, align=center] at (3*\u,-0.03) {coefficients\\[-2pt]$m\!\times\!A$};
\end{scope}
\node[align=left, anchor=north west, font=\tiny, text=cInk, text width=2.85cm] at (-0.2,-0.28)
  {For this token, the $A$ action fields share one orthonormal frame $Q$; $G$ says how much each axis uses each direction. $d{=}16$, $m{=}12$, $A{=}6$.};
\end{scope}
\draw[cTeach!85!black, line width=0.4pt, dashed] (tokNE) -- (cTL);
\draw[cTeach!85!black, line width=0.4pt, dashed] (tokSE) -- (cBL);
\begin{scope}[shift={(1.1,-2.55)}]
  \draw[cTeach, line width=1.0pt] (0,0) -- (0.24,0);
  \filldraw[fill=cTeach, draw=white, line width=0.3pt] (0,0) circle (1.4pt) (0.24,0) circle (1.4pt);
  \node[anchor=west, inner sep=2pt] at (0.32,0) {measured from $T$};
  \filldraw[fill=cData!90!black, draw=white, line width=0.3pt] (2.85,0) circle (1.4pt);
  \node[anchor=west, inner sep=2pt] at (2.95,0) {recorded data};
  \draw[cArith!85!black, line width=1.0pt, dashed] (4.95,0) -- (5.25,0);
  \node[anchor=west, inner sep=2pt] at (5.33,0) {computed in closed form};
\end{scope}
\end{tikzpicture}
\caption{\textbf{A chart of the action-conditioned data manifold.} \emph{(a)} For a fixed state $x$, varying the action traces a curved surface in latent space, the manifold $\mathcal{M}_x$, on which transitions recorded from $x$ lie. \emph{(b)} Probing the frozen generator $T$ at $\pm\delta$ (blue) gives the tangent plane at the current latent $z_t$. While $T$'s own response bends away from the manifold, the Action Jet (orange) fits the Taylor polynomial to the recorded transitions in the data and so follows $\mathcal{M}_x$ up to the radius $r_1^{*}$ (green) it computes from its own coefficients. \emph{(c)} One token of the chart is marked on the tangent plane in (b). The geometry encoder writes the fields of all $A$ action axes in one orthonormal frame $Q$ with coefficients $G$.}
\label{fig:teaser}
\end{figure}
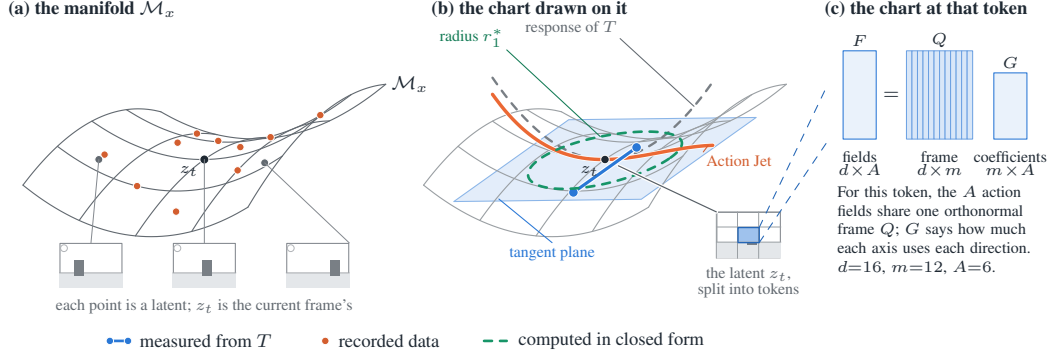

Planners over short horizons only require local information with respect to the current state. 
Fix the state $x$ of a scene and vary an action $a \in \mathbb{R}^{A}$, such as a camera displacement or a robot end-effector command, with one component per action axis.
The resulting latent image states trace a surface $\mathcal{M}_x$ (Fig.~\ref{fig:teaser}a), the action-conditioned data manifold. This manifold is a property of the data distribution, including the scenes, cameras and dynamics, and not of any specific model. Every action-conditioned generative model trained on that distribution carries an approximation of $\mathcal{M}_x$ that can be reached only by sampling.

This paper explicitly estimates the local shape of action-conditioned data manifolds.
A \emph{chart} identifies an open patch of an $n$-dimensional manifold with an open subset of $\mathbb{R}^n$, and we use the word for the map in the action-to-manifold direction, which is a chart where the $A$ action axes move the image independently. Freezing a generator $T$ and fixing the input state gives such a map from actions to latent future states, an approximate chart of $\mathcal{M}_x$ (Fig.~\ref{fig:teaser}b). We differentiate it with per-axis action probes to measure how each image region changes per unit of each action axis. 
The probe returns a field: one vector for every image token, a $2{\times}2$ block of the latent, saying how that token moves per unit of the action axis.
Stacked over the action axes, these fields form the \emph{action Jacobian}. A geometry encoder learns to emit that action Jacobian in one forward pass, factored as an orthonormal frame with per-axis coefficients (Fig.~\ref{fig:teaser}c).
A correction head adds the second- and third-order terms of the local Taylor expansion as coefficient fields of a polynomial in the action space, trained on recorded transitions (Fig.~\ref{fig:pipeline}). We compute from this Taylor polynomial the radius of validity for our approximated local shape or chart of the manifold. A \emph{jet} is an operation that takes a differentiable function and produces a Taylor polynomial at each point of its domain. We call our approximation method of the Taylor polynomial describing the local shape of the action-conditioned data manifold an \emph{Action Jet.}

\paragraph{Contributions and significance.} This paper introduces a method that estimates the local shape, specifically a chart, of action-conditioned data manifolds. By performing the local approximation of this manifold, we enable significant practical improvement for efficient and accurate candidate trajectory generation of image-action pairs. 
Instead of running the generative model for every action candidate, the Action Jet runs once per state, sampling the manifold, and every candidate within its radius of validity is then a contraction of stored coefficient fields via matrix multiplication alone.
Scoring sixteen candidates this way is over $300\times$ faster than with either source generative model, and the speed gap grows with the number of candidates.
Furthermore,
because only the tangent comes from the generator, and the data sets every higher-order term and corrects the tangent itself,
we can achieve more accurate future state predictions than the trained generator over short horizons.

\section{Related work}
\label{sec:related}

\noindent\textbf{Measuring the manifold hypothesis.} The hypothesis can be tested statistically \citep{fefferman2016testing} and studied more specifically for image data. The intrinsic dimension of image data has been estimated directly \citep{pope2021intrinsic,brown2023verifying}, from estimating pose-indexed appearance manifolds \citep{murase1995visual}, and from pretrained diffusion models \citep{stanczuk2024diffusion,kamkari2024geometric}.
Local polynomials have been shown to recover tangents and curvature from samples \citep{aamari2019nonasymptotic,philipona2003something}. We extend this research on manifold estimation to action-conditioned manifolds.

\noindent\textbf{Geometry of generative models.} Pullback metrics describe the curvature of a generator's latent space \citep{arvanitidis2018latent,shao2018riemannian,wang2021geometry}, and Jacobian and Hessian structure has been used to find or enforce interpretable directions
\citep{harkonen2020ganspace,peebles2020hessian,zhu2021lowrank,kwon2023diffusion,park2023understanding}.
These analyses concern one model's latent code; we differentiate with respect to a physical action and compare independently trained models.

\noindent\textbf{Jacobians and locally linear control.} Visual servoing relates the motion of an image to the motion of a camera through an image Jacobian \citep{hutchinson1996tutorial}, which can be estimated from observed motion without calibration and used inside a trust region \citep{jagersand1997experimental}, the classical counterpart of our probe and radius. Neural Jacobian Fields learn a dense first-order visuomotor Jacobian per robot \citep{li2025controlling}. Latent control models assume locally linear dynamics \citep{watter2015embed,banijamali2018robust} or
penalize curvature to keep them locally linear \citep{levine2020prediction}. Similarly to approaches learning image transformation operators \citep{rao1998learning,jayaraman2015learning}, we estimate the local shape of the manifold directly from the data for a computed radius where our estimate is valid. 

\noindent\textbf{Action-conditioned generators.} World models predict observations from a state and an action \citep{ha2018world,hafner2025mastering,hansen2024tdmpc2}, increasingly as video generators conditioned on robot actions \citep{yang2024learning,bruce2024genie,zhu2025irasim,quevedo2026worldgym,huang2026vid2world,guo2026ctrlworld,nvidia2025cosmos}
or as predictors in pretrained feature spaces
\citep{bar2025navigation,zhou2025dinowm,assran2025vjepa}. Camera-controlled video diffusion
\citep{wang2024motionctrl,he2025cameractrl,gao2024cat3d,yu2025viewcrafter,wu2026framecrafter} and feed-forward novel-view synthesis \citep{charatan2024pixelsplat,chen2024mvsplat,xu2025depthsplat,jin2025lvsm,wang2025vggt,szymanowicz2026lagernvs} play the same role for camera motion. Each model has to perform a forward inference pass for each candidate, and model performance is measured by the quality of predicted candidate trajectories \citep{kang2025far,motamed2026generative,duan2025worldscore,li2025worldmodelbench}. We use these models as instruments and compare their local derivatives, enabling all candidate predictions from each state to be calculated from the result of a single forward inference pass.

\noindent\textbf{Derivatives, distillation and agreement.} Training on derivatives goes back to Tangent Prop \citep{simard1991tangent} and the manifold tangent classifier \citep{rifai2011manifold}. Sobolev training \citep{czarnecki2017sobolev} and Jacobian matching \citep{srinivas2018knowledge} distill a teacher's derivatives along with its outputs. Output distillation \citep{hinton2015distilling,salimans2022progressive,song2023consistency,yin2024onestep,sauer2024adversarial} trains a student to reproduce its teacher, whose quality then bounds the student. We distill only the first derivative with respect to an action and fit the rest to the data itself. The agreement between independently trained models~\citep{kornblith2019similarity,bansal2021revisiting} has provided evidence that the models capture a shared reality \citep{huh2024platonic,kadkhodaie2024generalization}. We apply that reasoning to derivative fields that share a coordinate system by construction.

\section{Charting the action-conditioned data manifold}
\label{sec:method}

\begin{figure}[t]
\centering
\begin{tikzpicture}[fig, node distance=0.28cm,
  hdr/.style={font=\scriptsize\bfseries, anchor=west, inner sep=0pt},
  io/.style={align=center, inner sep=1pt, font=\scriptsize},
  cell/.style={minimum width=3.1cm, minimum height=0.68cm, font=\scriptsize},
  ]
\def\cw{3.52}
\node[hdr] at (0,1.62) {1\; probe the generator};
\node[hdr] at (\cw,1.62) {2\; chart the tangent space};
\node[hdr] at (2*\cw,1.62) {3\; fit higher orders to data};
\node[hdr] at (3*\cw,1.62) {4\; query};
\node[io]            (c1i) at (1.55,1.2)  {state $x_t$,\; actions $\pm\delta_i e_i$};
\node[frozen, cell]  (c1m) at (1.55,0.45) {frozen generator $T$,\\[-1pt] frozen encoder $E$};
\node[teach, cell]   (c1o) at (1.55,-0.55) {$\J_i=\dfrac{z^{+\delta_i}-z^{-\delta_i}}{2\delta_i}$};
\draw[flow] (c1i) -- (c1m); \draw[flow] (c1m) -- (c1o);
\node[io]            (c2i) at (\cw+1.55,1.2)  {latents $z_t,\ z_{t-1}$};
\node[teach, cell]   (c2m) at (\cw+1.55,0.45) {geometry encoder\\[-1pt](ViT, one pass)};
\node[teach, cell]   (c2o) at (\cw+1.55,-0.55) {$Q\in\R^{d\times m}$,\; $G\in\R^{m\times A}$\\[-1pt]field $F_i=QGe_i$};
\draw[flow] (c2i) -- (c2m); \draw[flow] (c2m) -- (c2o);
\node[io]            (c3i) at (2*\cw+1.55,1.2)  {noise $\varepsilon$ per token};
\node[data, cell]    (c3m) at (2*\cw+1.55,0.45) {correction head};
\node[data, cell]    (c3o) at (2*\cw+1.55,-0.55) {$L_i,\ H_{ij},\ C_{ijk}$\; deterministic\\[-1pt]$U^{(1)}_i(\varepsilon),\ U^{(2)}_{ij}(\varepsilon)$\; fill};
\draw[flow] (c3i) -- (c3m); \draw[flow] (c3m) -- (c3o);
\node[io]            (c4i) at (3*\cw+1.55,1.2)  {$K$ candidate actions $a$};
\node[arith, cell]   (c4m) at (3*\cw+1.55,0.45) {monomials of each $a$\\[-1pt]$\otimes$ stored coefficients};
\node[arith, cell]   (c4o) at (3*\cw+1.55,-0.55) {$\hat z = z_t + QGa + \tilde r(a,\varepsilon)$\\[-1pt]radius $r_1^{*}$ in closed form};
\draw[flow] (c4i) -- (c4m); \draw[flow] (c4m) -- (c4o);
\draw[sup, cTeach] (c1o.east) -- (c2o.west);
\node[io, font=\tiny, text=cData, anchor=west] (rec) at (2*\cw,-1.42) {Stage 2, fit to recorded transitions};
\draw[sup, cData] (c3o.south |- rec.north) -- (c3o.south);
\draw[flow, cMuted] (c2m.east) -- (c3m.west);
\draw[flow, cMuted] (c3o.east) -- (c4o.west);
\node[font=\tiny, text=cMuted, anchor=west] at (0,-1.42) {Stage 0, once per training state};
\node[font=\tiny, text=cTeach, anchor=west] at (\cw,-1.42) {Stage 1, fit to generator Jacobians};
\node[font=\tiny, text=cMuted, anchor=west] at (3*\cw,-1.42) {inference, 2 network passes per state};
\end{tikzpicture}
\caption{\textbf{Method overview.} Fill color marks what supervises each part: blue, the generator's measured Jacobians; orange, recorded transitions; green, no network at all. In column~1, $z^{\pm\delta_i} = \varphi_x(\pm\delta_i e_i)$ is the encoded render of each probe (Eq.~\ref{eq:probe}).}
\label{fig:pipeline}
\end{figure}
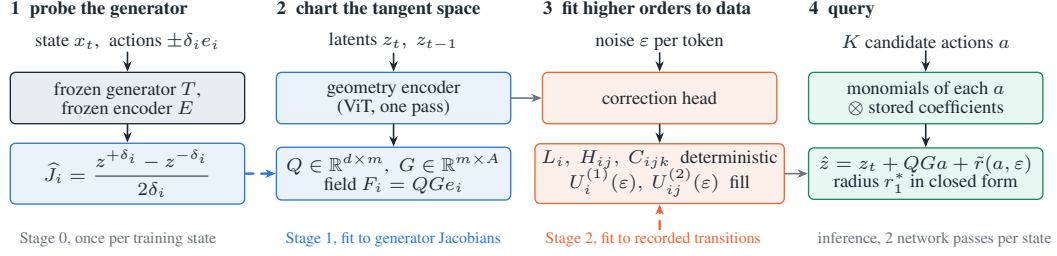

Our approach to estimating future states in the local region $\mathcal{M}_x$ has four steps corresponding to Sections~\ref{sec:probe}--\ref{sec:radius} which also correspond to each column of Figure~\ref{fig:pipeline}. First, we probe the generator to compute the action Jacobian for given input states (Section~\ref{sec:probe}). We use the results of this probe 
to train a geometry encoder to emit this action Jacobian directly, which gives the first-order term of the Taylor polynomial describing the shape of $\mathcal{M}_x$
(Section~\ref{sec:encoder}). We then fit the higher orders of the Taylor polynomial to the data itself (Section~\ref{sec:jet}). This Taylor polynomial is an approximation of $\mathcal{M}_x$ valid inside a radius within which we can query possible trajectory rollouts (Section~\ref{sec:radius}). 

\textbf{Problem Statement.} %
Latent image states $z$ are defined in the space of a frozen image autoencoder with encoder $E$ and decoder $D$ \citep{rombach2022ldm}: $z = E(x)$ for an image $x$. We split $z$ into $N$ tokens, each a $2{\times}2$ block of latent pixels, so every token is a vector in $\mathbb{R}^{d}$, with $d = 4c$ for a latent with $c$ channels ($d{=}16$ here), and $z \in \mathbb{R}^{N\times d}$. We call an element of $\mathbb{R}^{N\times d}$, one $d$-vector per token, a \emph{field}.
An action $a \in \mathbb{R}^{A}$ has one component $a_i$ per action axis, and $e_i$ is the $i$-th standard basis vector of $\mathbb{R}^{A}$. For an image state $x = x_t$ at time $t$, let $\mathcal{M}_x$ be the set of latent image states reachable from $x$ by actions in a neighborhood $\mathcal{U} \subset \mathbb{R}^{A}$ of zero.
This set is a smooth surface
of dimension at most $A$, fixed by the data distribution including dynamics.
Let $\psi_x : \mathcal{U} \to \mathcal{M}_x$ send an action to the latent state it reaches in the data (its mean, where transitions are stochastic), so that $\mathcal{M}_x = \psi_x(\mathcal{U})$ %
and $\psi_x(0) = z_t \coloneq E(x)$.
The local shape of $\mathcal{M}_x$ is the Taylor expansion
\begin{equation}
\psi_x(a) \;=\; z_t
+ \sum_i a_i J_i
+ \tfrac12 \sum_{i,j} a_i a_j H_{ij}
+ \tfrac16 \sum_{i,j,k} a_i a_j a_k C_{ijk} + \dots,
\label{eq:taylor}
\end{equation}

whose coefficients are fields, with indices running over the $A$ axes, such that the first-order fields $J_i = \partial\psi_x/\partial a_i$ span the tangent space, and $H_{ij}$ and $C_{ijk}$, symmetric in their indices, are the second- and third-order coefficients.
A generator $T$ trained on the dataset defines $\varphi_x(a) = E(T(x,a))$, and where it has learned the data, $\varphi_x \approx \psi_x$. We introduce \emph{Action Jets}, networks that emit a truncation of Eq.~\ref{eq:taylor} from the current frame, 
with the tangent measured from $\varphi_x$ and every other coefficient, including a correction to the tangent, fit to $\psi_x$ through recorded transitions 
(Fig.~\ref{fig:pipeline}).

\subsection{Probe the generator for the action Jacobian}
\label{sec:probe}

For each generator, state, and axis $i$ we render at $\pm\delta_i$ along that axis alone, encode both renders, and difference them into the field
\begin{equation}
\widehat{J}_i \;=\; \frac{\varphi_x(\delta_i e_i) - \varphi_x(-\delta_i e_i)}{2\delta_i}
\;\in\; \mathbb{R}^{N\times d}.
\label{eq:probe}
\end{equation}
$\widehat{J}_i$ estimates $\partial\varphi_x/\partial a_i$, the generator's version of $J_i$, and it says
how each token moves per unit of that axis. Since $\widehat{J}_i$ records the generator's response to the action rather than its rendering, the measurement is insensitive to how good the generator's samples look. For stochastic generators,
we average over multiple seed-paired samples (the same noise at $+\delta_i$ and $-\delta_i$) to measure
an expectation of $\widehat{J}_i$. The probe needs no gradients, so any model that maps a state and an action to an image can be probed, including multi-step diffusion samplers.

\subsection{Estimate the first-order Taylor polynomial with the geometry encoder}
\label{sec:encoder}

Probing costs two inference passes from the generator per axis per state, which a planner scoring many candidate actions cannot afford. We therefore train a \emph{geometry encoder} that predicts the action Jacobian from the current frame in one forward pass, supervised by the probed fields $\widehat J_i$ from Section~\ref{sec:probe}.

To construct this encoder, first a ViT reads $z_t$ and one context frame, the previous latent $z_{t-1} = E(x_{t-1})$,
as $N$ tokens. For each token $n$ it emits a feature vector $h_n$ (used by the correction head, Section~\ref{sec:jet}),
a matrix $B_n \in \mathbb{R}^{d\times m}$ where $m$ is the
rank of the shared tangent frame, and coefficients
$G_n \in \mathbb{R}^{m\times A}$. We orthonormalize $B_n$ by taking
$Q_n$ to be the $Q$ factor of the reduced QR decomposition of $B_n$, so %
$Q_n^{\top}Q_n = I_m$
by construction. 
We then predict how token $n$ moves per unit of axis $i$ as $F_{i,n} = Q_n G_n e_i \in \mathbb{R}^{d}$, and stack these over tokens into the field $F_i \in \mathbb{R}^{N\times d}$. Using this construction,
the geometry encoder minimizes
\begin{equation}
\mathcal{L}_{\mathrm{enc}}
= \sum_{i=1}^{A} \frac{\lVert F_i - \widehat{J}_i\rVert^{2}}{\lVert \widehat{J}_i\rVert^{2}}
\;+\; \lambda_{\mathrm{cov}} \sum_{i=1}^{A}
\frac{\lVert (I - Q Q^{\top})\widehat{J}_i\rVert}{\lVert \widehat{J}_i\rVert},
\label{eq:enc}
\end{equation}
where norms are Frobenius norms over all tokens, $(I-QQ^{\top})$ acts token-wise as $I - Q_nQ_n^{\top}$, and $\lambda_{\mathrm{cov}} > 0$ weights the coverage term.
Normalizing each axis by its own energy keeps axes that cause large image motion from dominating the gradient. The second term penalizes the part of each probed field that lies outside the predicted frame.

$F_i$ is the encoder's estimate of $J_i$, and we use this estimate to construct the first-order truncation of Eq.~\ref{eq:taylor}:
\begin{equation}
\widehat\psi^{(1)}_x(a)
\;=\; z_t + \sum_{i=1}^{A} a_i F_i .
\label{eq:firstorder}
\end{equation}

\subsection{Fit higher orders of the Taylor polynomial to the data}
\label{sec:jet}

The first-order chart of Eq.~\ref{eq:firstorder} is a plane and
$\mathcal{M}_x$ is a curved surface. We add the second- and third-order terms of Eq.~\ref{eq:taylor} 
with a \emph{correction head}, and truncating
at third order makes the prediction a \emph{3-jet}.
Unlike the geometry encoder, it is not supervised by the generator; it is trained on recorded transitions, so every term above first order is fit to the data itself.

For each token $n$ the correction head reads the encoder feature $h_n$ and a noise draw $\varepsilon_n\sim \mathcal{N}(0, I)$
and emits coefficient fields for the polynomial
\begin{equation}
r(a,\varepsilon) \;=\;
\sum_i a_i L_i
+ \tfrac12 \sum_{i,j} a_i a_j H_{ij}
+ \tfrac16 \sum_{i,j,k} a_i a_j a_k C_{ijk}
+ \sum_i a_i U^{(1)}_i(\varepsilon)
+ \sum_{i,j} a_i a_j U^{(2)}_{ij}(\varepsilon),
\label{eq:jet}
\end{equation}
where every field lies in $\mathbb{R}^{N\times d}$. $H_{ij}$ and $C_{ijk}$ are the head's estimates of the second- and third-order coefficients of Eq.~\ref{eq:taylor}, $L_i$ corrects the first-order term, and $U^{(1)}, U^{(2)}$
are noise-conditioned \emph{fill} fields for disoccluded content, with $\mathbb{E}_\varepsilon[U^{(2)}] = 0$ so that the mean quadratic term lives in $H$ (App.~\ref{app:formal}).
The prediction for action $a$ is
\begin{equation}
\hat z(a,\varepsilon) \;=\;
\widehat\psi^{(1)}_x(a)
+ \tilde r(a,\varepsilon)
\;=\; z_t + \sum_i a_i F_i + \tilde r(a,\varepsilon),
\label{eq:pred}
\end{equation}
with $\tilde r(a,\varepsilon) \coloneq r(a,\varepsilon) - r(0,\varepsilon)$. Every monomial in Eq.~\ref{eq:jet} vanishes at $a{=}0$, but the head's learned output bias does not, so subtracting $r(0,\varepsilon)$ removes it and guarantees $\hat z(0,\varepsilon) = z_t$ (App.~\ref{app:formal}). 
The correction head is trained on recorded transitions $(x_t, x_{t+g})$ with the encoder frozen, to predict $z_{t+g} = E(x_{t+g})$ under the recorded action, with latent, pixel $L_1$, LPIPS~\citep{zhang2018lpips} and conditional adversarial losses, and the adversarial critic is the one drawn in App.~Fig.~\ref{fig:pipeline_a}. A final stage fine-tunes the decoder $D$ on real frames with everything else frozen (``FT decoder'' in the tables). No generator is fine-tuned or run at inference.

\subsection{Query candidate future trajectories in the radius of validity}
\label{sec:radius}

Our estimate of the local manifold does not hold for actions associated with too large state transitions. Conceptually, this idea can be illustrated simply by camera movement. If a camera is moved to a place where there is no visual overlap with the current scene, it is impossible to predict the contents of the new view given the current view. Computationally, the Taylor polynomial estimating the local manifold shape is no longer accurate too far from its expansion point. Because Eq.~\ref{eq:jet} is an explicit polynomial in $a$, this \emph{radius of validity} can be computed from the chart's own coefficients.

Let $u$ be a unit direction in normalized action units. The noise-averaged displacement from $z_t$ at step length $r$ is exactly a cubic $P(r) \;=\; \mathbb{E}_\varepsilon[\hat z(ru,\varepsilon)] - z_t \;=\; \alpha r + \beta r^2 + \gamma r^3$ with $\alpha = \sum_i u_i \bigl(F_i + L_i + \mathbb{E}_\varepsilon[U^{(1)}_i]\bigr)$,
$\beta = \tfrac12 H[u,u] = \tfrac12\sum_{i,j} u_i u_j H_{ij}$ and %
$\gamma = \tfrac16 C[u,u,u] = \tfrac16\sum_{i,j,k} u_i u_j u_k C_{ijk}$. With the Frobenius norms $p_1 = \lVert\alpha\rVert$, $p_2 = \lVert\beta\rVert$ and $p_3 = \lVert\gamma\rVert$,
the \emph{radius of validity} along $u$ is $r_1^{*}(u) = \bigl(
-p_2 + \sqrt{p_2^2 + 4p_1p_3}\bigr) / (2p_3)$, the positive root of $p_2 r^2 + p_3 r^3 = p_1 r$, past which the nonlinear terms can be as large as the linear one. 
Inside $r_1^{*}$, any additional
candidate actions can be simulated from the current state. Outside that radius, a new chart needs to be approximated.

\section{Results}
\label{sec:results}

\begin{figure}[htbp]
  \nopagebreak
  \centering
  \begin{minipage}[b]{0.48\textwidth}
    \centering
    \includegraphics[width=\textwidth]{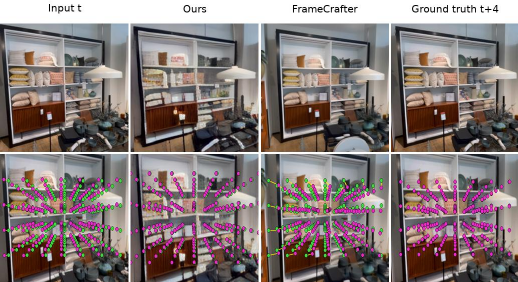}
    \caption{Top: Predictions on DL3DV Benchmark images at $g{=}4$. Bottom: points projected through the true (magenta) and recovered (green) camera poses, with their offset being the VCRE.}
    \label{fig:nvs_qual_paper_figure}
  \end{minipage}
  \hfill %
  \begin{minipage}[b]{0.48\textwidth}
    \centering
    \resizebox{\textwidth}{!}{
    \begin{tabular}{l cc cc }
    \toprule
    & \multicolumn{2}{c}{$g{=}1$} & \multicolumn{2}{c}{$g{=}4$}\\
    \cmidrule(lr){2-3}\cmidrule(lr){4-5}
    method & \makecell{rot.\\(deg)$\downarrow$} & \makecell{VCRE\\(px)$\downarrow$} & \makecell{rot.\\(deg)$\downarrow$} & \makecell{VCRE\\(px)$\downarrow$}\\
    \midrule
    \makecell[l]{GT frame vs.\ itself \\(instrument floor)} & 0.12 & 0.4 & 0.10 & 0.4\\
    \makecell[l]{Copy input frame \\(control)} & 0.39 & 3.7 & 1.73 & 19.1\\
    \midrule
    DepthSplat & 0.28 & 2.2 & 0.29 & 2.8\\
    LagerNVS & 0.19 & 1.3 & 0.27 & 1.8\\
    \midrule
    FrameCrafter & 0.89 & 8.0 & 2.16 & 51.1\\
    Ours & \textbf{0.26} & \textbf{1.6} & \textbf{0.79} & \textbf{5.9}\\
    \bottomrule
    \end{tabular}
    }

    \captionof{table}{Geometric consistency: median rotation error and VCRE (px) of the camera recovered between each prediction and the real frame at its target pose. Action Jets outperform the generative peer at a fraction of inference cost.}
    \label{tab:nvs_reprojection_metrics}
  \end{minipage}
\end{figure}

\subsection{Experimental setup}
\label{sec:setup}

We evaluate Action Jets on three different domains. \emph{Kubric} synthetic scenes \citep{greff2022kubric} use three action axes (camera azimuth, object translation, light azimuth) and a differentiable generator, so every probe can be checked against the generator's exact Jacobian.
\emph{BridgeData~V2} robot manipulation \citep{walke2023bridgedata} has two
independently trained action-conditioned world models, IRASim
\citep{zhu2025irasim} and WorldGym \citep{quevedo2026worldgym}, and we use it to measure where the learned chart holds and its downstream utility.
\emph{DL3DV} \citep{ling2024dl3dv}, showing camera motion in static environments, has two architecturally unrelated novel-view generators, DepthSplat, built on explicit
Gaussian splatting \citep{xu2025depthsplat} with depth estimation, and LagerNVS, a transformer initialized
from VGGT \citep{szymanowicz2026lagernvs,wang2025vggt}. We also use FrameCrafter \citep{wu2026framecrafter} as a diffusion-based generative baseline.  DL3DV is used to show that the extracted object belongs to the data rather than the models we probed.

\subsection{Kubric: certifying the instrument}
\label{sec:kubric}

Kubric is the only setting with three declared generative factors (camera azimuth, object translation, light azimuth) and a differentiable generator: we train a feed-forward next-frame transformer on these scenes, so the quantity Eq.~\ref{eq:probe} estimates is available exactly by automatic differentiation.

Probe-extracted fields match the generator's autodiff Jacobian at
cosine $0.87/0.99/0.90$ per factor at $\delta{=}0.02$, and the encoder
reaches held-out per-axis cosines $0.73/0.67/0.77$.
All three axes share one rank-$m$ frame (\S\ref{sec:encoder}), and the effective rank of the target energy
in that frame is $3.9$ of $m{=}12$, against three declared factors.

Table~\ref{tab:kubric} evaluates the 3-jet architecture (\S\ref{sec:jet}) against tangent-only prediction, 
and an unconstrained Transformer head that runs every query,
and so forfeits exactly the property the jet exists to preserve. The free-form Transformer head buys $1.0$--$2.6$\,dB, shrinking with the gap, and is at LPIPS parity with the 3-jet by gap~3. The jet's advantage over the tangent-only chart, by contrast, grows with the gap, from $+1.4$ to $+3.5$\,dB, which is the division of
labor of \S\ref{sec:jet} showing that the further the step, the more of the displacement lives above first order. Per-factor disentanglement, the holonomy probe, and the ablation that shows sharpening a generator improves its renders and corrupts its derivatives are in App.~\ref{app:kubric_extra}.

\begin{table}[t]
\centering\small
\begin{tabular}{lcccc|cccc}
\toprule
& \multicolumn{4}{c|}{PSNR $\uparrow$ (gap 1/2/3/4)} &
\multicolumn{4}{c}{LPIPS $\downarrow$ (gap 1/2/3/4)} \\
\midrule
tangent only & 30.43 & 26.48 & 23.66 & 22.80 & .0241 & .0486 &
.0720 & .0926 \\
free-form Transformer head & 34.45 & 31.19 & 28.28 & 27.26 & .0153 &
.0267 & .0391 & .0511 \\
3-jet correction head (ours) & 31.80 & 29.05 & 26.92 & 26.27 & .0183 &
.0305 & .0403 & .0501 \\
\bottomrule
\end{tabular}
\caption{Kubric $4$-frame horizon prediction through the shared fine-tuned decoder. The free-form head re-runs a Transformer network for every candidate action, while the 3-jet answers any candidate by arithmetic on one extraction. The 3-jet reaches LPIPS parity at gaps 3--4 and 
trails by $1.0$--$2.6$\,dB in PSNR, a gap that shrinks with the horizon. The 3-jet dominates the 2-jet at matched decoder,
and decoder fine-tuning is a strong rendering lever applied to all variants
(App.~Table~\ref{tab:kubric_ablation}).}
\label{tab:kubric}
\end{table}

\subsection{BridgeData~V2: where the chart holds and its utility}
\label{sec:bridge}

BridgeData~V2 is a table-top manipulation dataset where a 7-DoF arm moves
against a mostly static scene. Two independently trained action-conditioned world models cover it, IRASim and WorldGym, each probed in its own action convention, and we learn a chart from each.

The two generators turn out to define almost unrelated tangent
structures on this dataset where per-axis 
$\cos(\widehat J_A, \widehat J_B) = -0.04$
to $0.05$, against a ceiling of at least $0.25$--$0.67$ that two
generators sharing a signal would have to reach, so the disagreement is
not attenuation by sampling noise, and the charts inherit it
($\cos(F_A, F_B) = -0.09$ to $0.12$). App.~\ref{app:agree_bridge} gives the reliability bounds, the convention-invariant subspace comparison, and the held-out anchoring test that 
identifies the chart learned from IRASim as the one that belongs to the data, which is the chart we evaluate. The
WorldGym-basis chart is kept as a negative control and as a maximally dissimilar chart 
on which to repeat the radius measurement.

\paragraph{Open-loop prediction and policy consistency.}
We roll out $441$ held-out validation episodes under the recorded
actions and score each predicted frame against the real one at gaps
$1$--$4$ (Table~\ref{tab:bridge_vis_metrics}). Action Jet with its
fine-tuned decoder beats IRASim, the generator that supplied its
tangent, at gap~1 on PSNR and LPIPS and is level at gap~2, 
at over $100\times$ less compute ($37.9$ against $5478$\,ms per state, Table~\ref{tab:bridge_nfe}). From gap~3 IRASim leads by $0.6$--$1.0$\,dB, the benefit of video diffusion denoising the whole trajectory jointly.
We also use the Octo policy \citep{octo2024} to
measure policy consistency as ADE and FDE between the action
trajectories Octo returns on predicted and on real gap-4 futures under
paired sampling keys, so that a perfect simulator scores zero. App.~\ref{app:metric} explains the metric in more detail. This is
the one axis where IRASim consistently stays ahead ($0.846$ against $1.299$).

\begin{table}[t]
\centering\footnotesize
\setlength{\tabcolsep}{3.6pt}
\begin{tabular}{l cccc cccc cc}
\toprule
& \multicolumn{4}{c}{PSNR $\uparrow$} & \multicolumn{4}{c}{LPIPS $\downarrow$} & \multicolumn{2}{c}{policy} \\
\cmidrule(lr){2-5}\cmidrule(lr){6-9}\cmidrule(lr){10-11}
gap $g$ & 1 & 2 & 3 & 4 & 1 & 2 & 3 & 4 & ADE $\downarrow$ & FDE $\downarrow$ \\
\midrule
copy-frame (control) & 24.48 & 21.48 & 20.04 & 19.19 & \textbf{.051} & \textbf{.084} & \textbf{.105} & \textbf{.121} & 1.577 & 1.994 \\
WorldGym & 23.80 & 21.10 & 19.70 & 18.86 & .059 & .093 & .118 & .137 & 1.448 & 1.783 \\
IRASim (source generator) & 23.34 & 22.37 & \textbf{21.64} & \textbf{21.12} & .101 & .116 & .127 & .136 & \textbf{0.846} & \textbf{1.035} \\
Action Jet & 24.21 & 22.07 & 20.67 & 19.67 & .074 & .110 & .144 & .177 & 1.474 & 1.992 \\
\;\;$+$ fine-tuned decoder & \textbf{24.57} & \textbf{22.42} & 21.07 & 20.11 & .073 & .111 & .143 & .170 & 1.299 & 1.663 \\
\bottomrule
\end{tabular}
\caption{Open-loop rollouts on BridgeData~V2 ($n{=}441$), and Octo policy consistency ($H{=}4$, a perfect simulator scores $0$). The Action Jet basis is charted from IRASim. Action Jet leads on short horizons against its generator.}
\label{tab:bridge_vis_metrics}
\end{table}

\begin{figure}[htbp]
  \nopagebreak
  \centering
  \begin{minipage}[c]{0.66\textwidth}
    \centering
    \includegraphics[width=\textwidth]{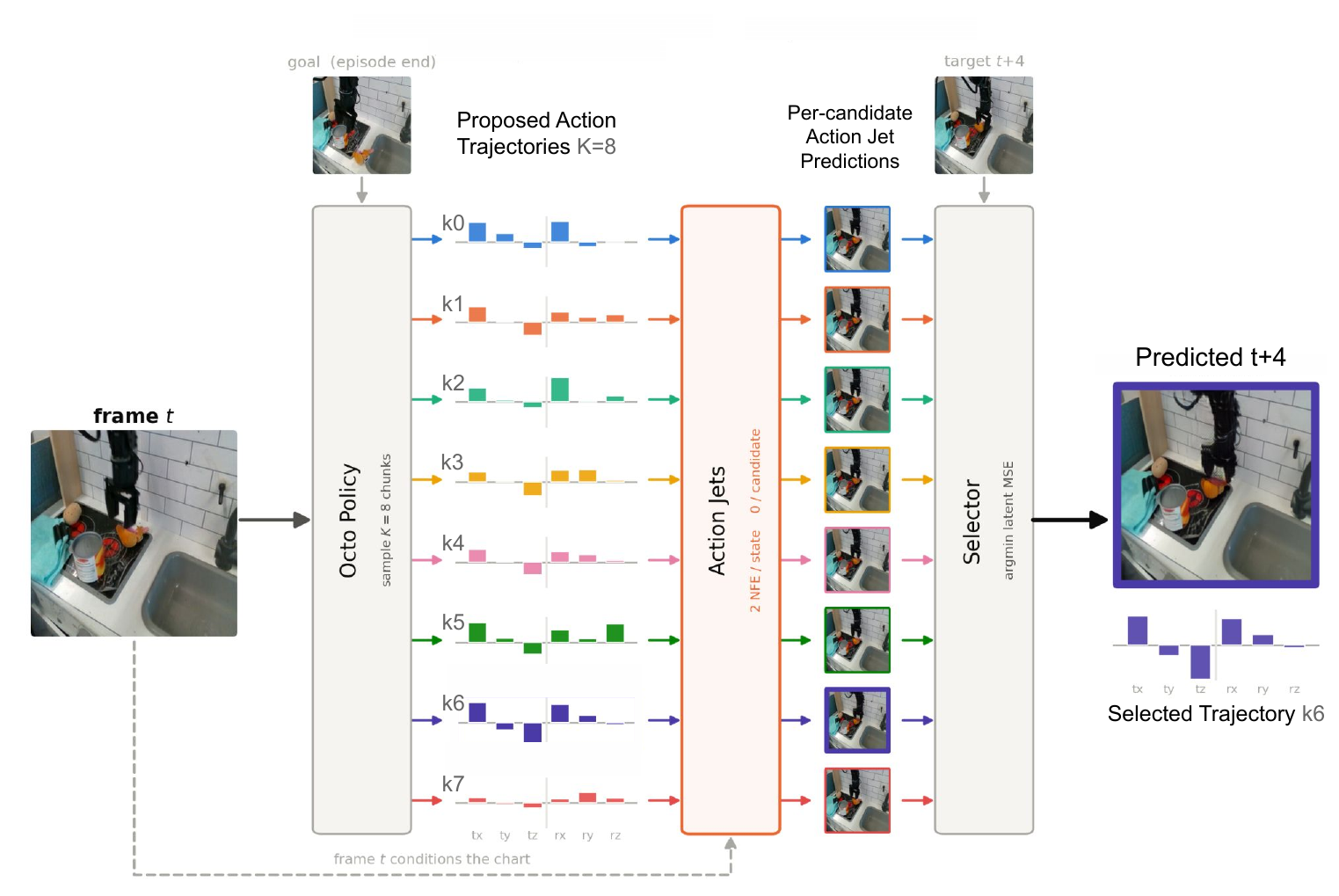}
    \caption{\textbf{Parallel Octo policy candidate prediction and evaluation.} 
    Candidate trajectories sampled from Octo are predicted by the Action Jet, and the winning candidate is the one with the lowest latent MSE to the true $t{+}4$ observation. Octo takes the goal image as its target and proposes candidate trajectories four steps into the future at each planning step, and the planning loop repeats until the goal state is reached.}
    \label{fig:fanout}
  \end{minipage}
  \hfill
  \begin{minipage}[c]{0.32\textwidth}
    \centering
    \small
    \setlength{\tabcolsep}{2.5pt}
    \resizebox{\textwidth}{!}{
        \begin{tabular}{l rrrr}
        \toprule
        \multicolumn{5}{l}{\emph{cost}}\\
        verifier & \multicolumn{2}{r}{ms/cand $\downarrow$} & \multicolumn{2}{r}{$K$@1s $\uparrow$}\\
        \midrule
        IRASim & \multicolumn{2}{r}{5714} & \multicolumn{2}{r}{0.2}\\
        WorldGym & \multicolumn{2}{r}{1094} & \multicolumn{2}{r}{0.9}\\
        \makecell[l]{Ours} & \multicolumn{2}{r}{\textbf{36.7}} & \multicolumn{2}{r}{\textbf{27}}\\
        \midrule
        \multicolumn{5}{l}{\emph{selected-action MSE $\downarrow$ at $K$}}\\
        \midrule
        verifier & 4 & 16 & \multicolumn{2}{r}{32} \\
        \midrule
        \makecell[l]{random-of-$K$\\(control)} & .040 & .038 & \multicolumn{2}{r}{.038} \\
        \makecell[l]{oracle-of-$K$\\(ceiling)} & .006 & .003 & \multicolumn{2}{r}{.003} \\
        \addlinespace[2pt]
        IRASim & .031 & .029 & \multicolumn{2}{r}{.030} \\
        WorldGym & .044 & .056 & \multicolumn{2}{r}{.063} \\
        \makecell[l]{Ours} & \textbf{.030} & \textbf{.028} & \multicolumn{2}{r}{\textbf{.025}} \\
        \bottomrule
        \end{tabular}
    }
    \vspace{2pt}
    \captionof{table}{Goal-image candidate verification under a compute budget. $K$@1s is the number of candidates affordable per second, where Action Jet is the only viable verifier among all baselines. Action Jet is at parity on smaller number of candidates and outperforms all baselines as the number of candidates increases.}
    \label{tab:bridge_budget}
  \end{minipage}
\end{figure}

\paragraph{Cost.}
\label{sec:speed}
For most world models, screening each action trajectory candidate costs a full forward pass. Table~\ref{tab:bridge_nfe} measures the time to evaluate $K$ candidate four-step futures at one state on identical hardware. Going from one candidate to sixteen multiplies the two diffusion world models' inference time by $16.7$ and $14.1$. Action Jet costs two network evaluations per state regardless of $K$, so sixteen candidates cost $1.4\times$ one and sixty-four cost $3.1\times$, 
so the average cost per candidate falls to $1.8$\,ms at $K{=}64$ (the marginal one costs $1.3$\,ms) and decreases with breadth.
Ranking in latent space means the decoder runs once rather than $K$ times.
Fig.~\ref{fig:fanout} shows eight candidates on a held-out state, all from a single chart extraction.

\begin{table}[t]
\centering\footnotesize
\setlength{\tabcolsep}{5pt}
\begin{minipage}[t]{0.51\textwidth}
\centering
\setlength{\tabcolsep}{2.2pt}
\begin{tabular}{l r rrr rr}
\toprule
& & \multicolumn{3}{c}{ms/state at $K$} & ms/ & $K$@ \\
\cmidrule(lr){3-5}
model & NFE & $1$ & $16$ & $64$ & cand & 1s$\uparrow$ \\
\midrule
IRASim & $25K$ & 5478 & 91431 & --- & 5714 & 0.2 \\
WorldGym & $40K$ & 1240 & 17501 & --- & 1094 & 0.9 \\
\midrule
AJ, composed & $8$ & 87.8 & 604.3 & 2347.9 & 36.7 & 27 \\
AJ, single & $\mathbf{2}$ & \textbf{37.9} & \textbf{54.0} & %
\textbf{117.5} & \textbf{1.8} & \textbf{545} \\
\bottomrule
\end{tabular}
\captionof{table}{Measured cost in milliseconds of evaluating $K$ candidate four-step futures at one state, where ms/cand is ms/state divided by $K$ at the largest $K$ measured, and $K$@1s is $1000$ over ms/cand. Action Jet, even when rolled out autoregressively via composed rollout, is orders of magnitude faster than both diffusion baselines.}
\label{tab:bridge_nfe}
\end{minipage}\hfill
\begin{minipage}[t]{0.47\textwidth}
\centering
\setlength{\tabcolsep}{2.2pt}
\begin{tabular}{lcc}
\toprule
 & IRASim & WorldGym \\
\midrule
copy parity, $|a|/r_1^{*}$ & 1.89 & 2.04 \\
peak gain vs.\ copy & $+.053$ @$1.02$ & $+.053$ @$1.18$ \\
1st-order/jet err. & $1.13 \to 0.80$ & $1.10 \to 0.70$ \\
rank corr.\ $(r_1^{*}, r_{\mathrm{emp}})$
& $+0.61$ {\tiny($n{=}28$)} & $+0.55$ {\tiny($n{=}29$)} \\
median $r_{\mathrm{emp}}/r_1^{*}$ & 1.72 & 2.02 \\
motion realized & 0.71 & 0.63 \\
\bottomrule
\end{tabular}

\captionof{table}{The radius against real Bridge frames for two charts, each in its \emph{own} $r_1^{*}$ units (96 states, gaps $1$--$12$). $r_{\mathrm{emp}}$ is the step at which the chart stops beating the copy-frame control. ``Motion realized'', $\norm{\hat z(a) - z_t}/\norm{z_{t+g}-z_t}$, rules out the stationary-frame degeneracy discussed in App.~\ref{app:radius_def}.}
\label{tab:two-charts}
\end{minipage}
\end{table}

\paragraph{Breadth against fidelity.}
\label{sec:beyond}
Planning utility is fidelity times the breadth of futures a planner can
afford to weigh. Each world model acts as a goal-image verifier over a pool of $K$ candidate action chunks sampled from Octo, predicting every outcome, ranking by latent similarity to the real $t{+}4$ frame, and judging the selected candidate trajectory, with random-of-$K$ and oracle-of-$K$ bracketing every curve (Table~\ref{tab:bridge_budget}).
Breadth helps competent verification, where the oracle and our chart improve with pool size, IRASim holds level, and the oracle-of-$K$ ceiling halves from $K{=}4$ to $16$, which shows the benefit of breadth search.
Verification skill dissociates from
simulation fidelity, shown by WorldGym anti-selecting with its choices worsening from $0.044$ to $0.063$ as the pool grows, despite mid-pack pixel metrics in Table~\ref{tab:bridge_vis_metrics}. And under a fixed budget the ranking inverts, since one second buys IRASim a fifth of a candidate while Action Jet evaluates the entire pool. Action Jet wins on both speed and verification.

\paragraph{Validity radius verification.}
\label{sec:radius_emp}
\S\ref{sec:radius} computes $r_1^{*}$ from the correction head's own
coefficients, and  Bridge lets us check it against real frames since the action from frame $i$ to frame $i{+}g$ grows with the gap, with corresponding ground truth frames available. We sweep over $g = 1 \ldots 12$ horizons, express each step as $|a| / r_1^{*}(a/|a|)$, which is the demonstrated action in units of the radius the Action Jet predicts along that same direction, and score the decoded prediction in LPIPS against the real frame, alongside the copy baseline, which scores frame $i$ itself
against frame $i{+}g$. Measuring the validity radius coverage, we find that $99\%$ of single-step transitions fall inside $r_1^{*}$ and $67\%$ of four-step ones, the horizon the correction head was trained on. The chart beats the copy baseline out to $1.89\,r_1^{*}$ and the latent-space error ratio inverts at the same place (Table~\ref{tab:two-charts}), so a pixel metric and a latent one agree on where the expansion stops. The curves and the per-gap coverage for both charts and the per-state calibration are in App.~\ref{app:radius} (Figs.~\ref{fig:bridge-curve} and~\ref{fig:bridge-wg}, Table~\ref{tab:bridge-coverage}).

Because $r_1^{*}$ is computed from a chart alone, we can repeat
the same measurement on the WorldGym chart, which shows nearly orthogonal tangent fields, translation gains a factor of ten apart, a different action convention, and no anchoring on real transitions at all. Both charts nonetheless cross the copy baseline at $1.89$ and $2.04$ times their own radius, peak at the same advantage, invert the same error ratio, and rank states positively (Table~\ref{tab:two-charts}). This supports our findings about the validity radius, where $r_1^{*}$ is computed from the Action Jet alone, at inference cost, and predicts where that Action Jet stops beating the do-nothing baseline, to within a factor stable across two charts that agree about nothing else. Full curves and coverage by gap are in App.~\ref{app:radius}.

\subsection{DL3DV: two generators, one manifold}
\label{sec:dl3dv}

DL3DV is a dataset captured by a moving camera in static indoor and outdoor environments, 
and the action axes are the camera's 3D translation and rotation. We use two architecturally distinct generators: DepthSplat, which uses explicit Gaussian splatting on estimated depth, and LagerNVS~\citep{szymanowicz2026lagernvs}, a transformer initialized from VGGT~\citep{wang2025vggt}.
The hypothesis is that the dataset manifold chart that we learn should be generator-independent.

\paragraph{Forward prediction.} Table~\ref{tab:nvs_visual_metrics}
scores forward prediction, where each baseline is provided with two trailing context frames, against the ground truth target $t{+}g$, on the central 256-pixel window every renderer covers. Among generative methods ours leads FrameCrafter on PSNR and LPIPS at every gap, 
with every paired 95\% CI on the LPIPS difference excluding zero, from $[+0.029, +0.064]$ at $g{=}1$ to $[+0.157, +0.222]$ at $g{=}4$ (App.~Table~\ref{tab:nvs_all}). Reconstruction methods re-render observed pixels from posed views,
so we report them as the bound they are and not as competitors. Against the copy-frame control ours wins on PSNR
everywhere and on LPIPS at $g \in \{3,4\}$, ties at $g{=}2$ and loses only at $g{=}1$.

\begin{table}[t]
\centering\footnotesize
\begin{minipage}[t]{0.51\textwidth}
\centering
\setlength{\tabcolsep}{1.4pt}
\begin{tabular}{l cccc cccc}
\toprule
& \multicolumn{4}{c}{PSNR$\uparrow$} & \multicolumn{4}{c}{LPIPS$\downarrow$}\\
\cmidrule(lr){2-5}\cmidrule(lr){6-9}
gap $g$ & 1 & 2 & 3 & 4 & 1 & 2 & 3 & 4\\
\midrule
copy-frame & 18.6 & 16.0 & 14.8 & 14.1 & .122 & .205 & .267 & .315\\
\midrule
\multicolumn{9}{l}{\emph{reconstruction (a bound)}}\\
DepthSplat & 25.6 & 23.2 & 21.1 & 19.5 & .047 & .075 & .104 & .133\\
LagerNVS & 30.5 & 27.4 & 25.2 & 23.7 & .039 & .062 & .088 & .115\\
\midrule
\multicolumn{9}{l}{\emph{generative}}\\
FrameCrafter & 17.4 & 15.4 & 14.3 & 13.7 & .187 & .296 & .366 & .413\\
Ours & \textbf{20.4} & \textbf{18.7} & \textbf{17.5} & \textbf{16.8} & \textbf{.162} & \textbf{.207} & \textbf{.249} & \textbf{.283} \\
\bottomrule
\end{tabular}
\captionof{table}{DL3DV forward prediction, baselines predicting from the two trailing context frames, compared against the ground truth target $t{+}g$, scored on the central 256-pixel window. A chart learned from the other generator (DepthSplat) matches this one (learned from LagerNVS) to within $0.006$ LPIPS and $0.09$\,dB at every gap (App.~Table~\ref{tab:nvs_all}).}
\label{tab:nvs_visual_metrics}
\end{minipage}\hfill
\begin{minipage}[t]{0.475\textwidth}
\centering
\setlength{\tabcolsep}{1.6pt}
\begin{tabular}{lcccccc|c}
\toprule
 & $t_x$ & $t_y$ & $t_z$ & $r_x$ & $r_y$ & $r_z$ & mean \\
\midrule
\multicolumn{8}{l}{\emph{DL3DV, $n{=}256$, $\rho_A{=}\rho_B{=}1.000$}}\\
learned bases & .77 & .81 & .77 & .84 & .82 & .79 & \textbf{.80}\\
generators & .72 & .73 & .68 & .69 & .84 & .86 & .75\\
perm.\ null & .00 & .00 & .00 & .00 & .00 & .00 & .00\\
\midrule
\multicolumn{8}{l}{\emph{Bridge, $n{=}148$}}\\
learned bases & $-$.02 & .06 & .08 & $-$.09 & .09 & .12 & \textbf{.04}\\
generators & $-$.02 & .03 & .05 & $-$.04 & .04 & .05 & .02\\
ceiling $\ge$ & .65 & .67 & .59 & .39 & .44 & .25 & .50\\
\bottomrule
\end{tabular}

\captionof{table}{Tangent agreement $\cos(F_A,F_B)$ of learned bases and 
$\cos(\widehat J_A,\widehat J_B)$ of generators (median, at $0.25\,r_1^{*}$; $\rho$ is generator reliability, App.~\ref{app:agree_noise}).
Learned bases match and even exceed their generators' agreement where it exists, or inherit its absence (App.~\ref{app:agree_bridge}).}
\label{tab:agreement}
\end{minipage}
\end{table}

\paragraph{Geometric consistency.} Pixel metrics often cannot distinguish a
frame that moved the camera correctly from one that blurred toward the
mean, so we score the geometry instead  (Table~\ref{tab:nvs_reprojection_metrics}, Fig.~\ref{fig:nvs_qual_paper_figure}). We pair each prediction with the
real frame at the same target pose, match with MASt3R \citep{leroy2024mast3r}, recover the relative camera by PnP, and note that its ground truth is the identity. A frame against itself gives the instrument's noise ($0.12^{\circ}$, $0.4$\,px), and the copy control scores the camera motion a predict-nothing model leaves uncorrected. At $g{=}4$ Action Jet cuts the control's rotation error 
by $54\%$ and its virtual-correspondence reprojection error (VCRE) by $69\%$, and is below the control on $97\%$ of states, and beats our generative baseline FrameCrafter.
We also provide generators' metrics as the bound and not as competitors.

\paragraph{Two charts, built from unrelated generators, agree.}
\label{sec:evidence}
We probe each generator at $0.25$ of each axis's own $r_1^{*}$ on 256 identical held-out states, chart a basis from each, and compare the resulting tangent fields $F = QG$ axis by axis. Each geometry encoder is supervised only by its generator's probes, so $F$ never sees a real future frame.
The two charts agree at $0.77$--$0.84$ per axis, mean $0.80$, more than
their generators, which agree at mean $0.75$, as shown in Table~\ref{tab:agreement}. On four of six axes the learned bases agree with each other more than their generators do,
which is what two estimators that each discard the part of their supervision the other does not share would produce, and at a $1\sigma$ probe the two are level (App.~\ref{app:agree_nvs}).
The tangent subspaces coincide at mean principal cosine $0.74$ and the induced metrics agree to a relative Frobenius difference of $0.089$, so the two charts assign the same lengths as well as the same
directions (App.~\ref{app:agree_nvs}). Between BridgeData~V2 and DL3DV, the architecture, latent space and downstream real-data supervision are held fixed across the two domains, only the generators' agreement changes, and the bases' agreement changes with it. Thus we empirically validate the hypothesis of charts learning the underlying structure of the action-conditioned data manifold.

\section{Conclusion}

Action Jets treat the local shape of an action-conditioned data manifold as something to be measured rather than modeled. A single extraction scores any number of candidates in milliseconds, a capability no generative baseline can match, and the fitted curvature beats its source generator at short horizons. Each chart states its validity radius in closed form, and charts built from unrelated generators recover the same tangent structure, evidence that what we extract
belongs to the data rather than to one model.

\textbf{Limitations.} All geometry is measured in one fixed autoencoder's latent space, so strictly we recover the composite data-and-encoder manifold, and our experiments vary the generator, not the latent. Disocclusion has no polynomial form, so even with noise injection there is some blur that concentrates at visibility boundaries. Each chart holds only at its probe scale. The advantage over the source generator holds at short horizons. All evaluations are open-loop. 

\subsection*{Ethics statement}
This work studies pretrained generative models and public datasets of robot manipulation and indoor scenes. It involves no human subjects, no personally identifying data and no new data collection. The measurements we describe are intended to make the failure modes of action-conditioned generators easier to detect before deployment, and the negative result we report about one publicly described world model is offered in that spirit.

\subsection*{Reproducibility statement}
Appendix~\ref{app:results} gives the probing protocol in full (App.~\ref{app:agree} and~\ref{app:controls_guide}), including window derivation,
state selection, noise averaging and the reliability correction.
Appendix \ref{app:formal} states the estimator properties and identifiability conditions with derivations. Every agreement in the paper is reported with its floor, and every stochastic comparison with its ceiling.

\subsection*{AI use statement}
Generative AI tools were used for manuscript drafting and copy-editing, for producing explanatory diagrams from the authors' own figure specifications, and for
searching and verifying related-work citations, as well as polishing code.
They were not used for research ideation, experimental design, or analysis; the probing harness,
models, experiments and all reported numbers are the authors' own work. The authors reviewed all AI-assisted text and figures, verified every number against the
experiment logs that produced it, and take responsibility for the final content of this work.

\bibliographystyle{iclr2027_conference}
\bibliography{references}

\newpage
\appendix
\section{Formal statements}
\label{app:formal}

\paragraph{Setup.} Fix a frozen autoencoder $(E,D)$ and a state $x$ with latent
$z = E(x)$, processed as $N$ blocks so that predictions live in
$\mathcal{Z} = \R^{N\times d}$.
A conditional generator $T$ induces the chart
map $\varphi_x : U \subset \R^{A} \to \mathcal{Z}$,
$\varphi_x(a) = E(T(x,a))$, assumed $C^3$ on a neighborhood of $0$ away from
visibility events. 
To the extent $T$ has learned the data, $\varphi_x$
approximates the data map $\psi_x$ of Eq.~\ref{eq:taylor}, whose
tangent fields are $J_i = \partial\psi_x/\partial a_i|_0$.

\paragraph{Probe accuracy.} Taylor expansion with the even orders canceling
gives $\J_i = \partial_i\varphi_x + \tfrac{\delta^2}{6}\partial^3_{iii}\varphi_x + O(\delta^4)$ at $a{=}0$, an estimate of $J_i$ to the extent $\varphi_x \approx \psi_x$,
second-order accurate in $\delta$ with all even-order terms removed. For a
stochastic generator $T(x,a,\xi)$ the probe estimates $\E_\xi[\J_i]$, and
averaging $M$ seed-paired renders reduces the noise variance by $1/M$.

\paragraph{Reliability ceiling.} Write a measured field as $\J = J + \eta$ with
noise of relative power $1-\rho$, where $\rho$ is the split-half reliability
estimated from seed-paired repetitions and Spearman--Brown corrected to the $M$-average.
For two generators with reliabilities $\rho_A, \rho_B$ and true-field cosine $c = \cos(s_A, s_B)$,
$\E[\cos(\J_A,\J_B)] \approx c\sqrt{\rho_A \rho_B}$, and with one deterministic generator the ceiling is $\sqrt{\rho_B}$.

\paragraph{Curvature and the ladder.}
Expanding $\varphi_x$ to third order as Eq.~\ref{eq:taylor} expands $\psi_x$, per-unit fields
probed at scales $\delta_1 < \delta_2$ along axis $i$ differ by $\J_i(\delta_2) - \J_i(\delta_1) 
= \tfrac{\delta_2^2 - \delta_1^2}{6}\,\partial^3_{iii}\varphi_x + \dots$,
so the measured decorrelation of per-unit fields across probe scales is a direct, axis-resolved image of higher-order structure.

\paragraph{Parity and identifiability.}
With the tangent supplied by the encoder, 
the residual $e(a) = \psi_x(a) - z_t - QGa$
expands with an odd deterministic leading term $(J-QG)a$. The linear fields $L_i$ of the correction head are the parity home of that term; omitting them forces the load through the noise-conditioned fill, where it arrives jittered across hypotheses. Deterministic and stochastic quadratic fields are separated by $\E_\varepsilon[U^{(2)}(\varepsilon)] = 0$, since any mean is absorbed into $H$, so $(H, U^{(2)})$ are identifiable in expectation.

\paragraph{The identity gauge.}
Every monomial in $r(a,\varepsilon)$ (Eq.~\ref{eq:jet}) carries at least one factor of $a$, so the correction head's polynomial satisfies $r(0,\varepsilon) = 0$; a learned readout bias $b$ violates it.
The gauged head $\tilde r = r(a,\varepsilon) - r(0,\varepsilon)$ removes any $a$-independent term exactly while acting as the identity on all monomial terms, hence $\tilde r$ equals the polynomial of Eq.~\ref{eq:jet} whenever the leak is
additive.

\paragraph{Single-inference complexity.} For $A{=}6$: quadratic monomials
$n_2 = \binom{A+1}{2} = 21$ and cubic monomials $n_3 = \binom{A+2}{3} = 56$.
Scoring $K$ candidate actions at one state costs exactly two network evaluations,
plus one fill pass per additional noise hypothesis, and
multiply--adds; no decode is required for
latent-space screening.

\section{Extended results}
\label{app:results}
\label{app:extra}

This supplement carries the measurements quoted in \S\ref{sec:results}. The instrument ladders and the control inventory that the rest of the supplement refers to are collected at the end, in App.~\ref{app:controls_guide}.

\subsection{The method end to end}

\begin{figure}[h]
\centering
\includegraphics[width=0.95\textwidth]{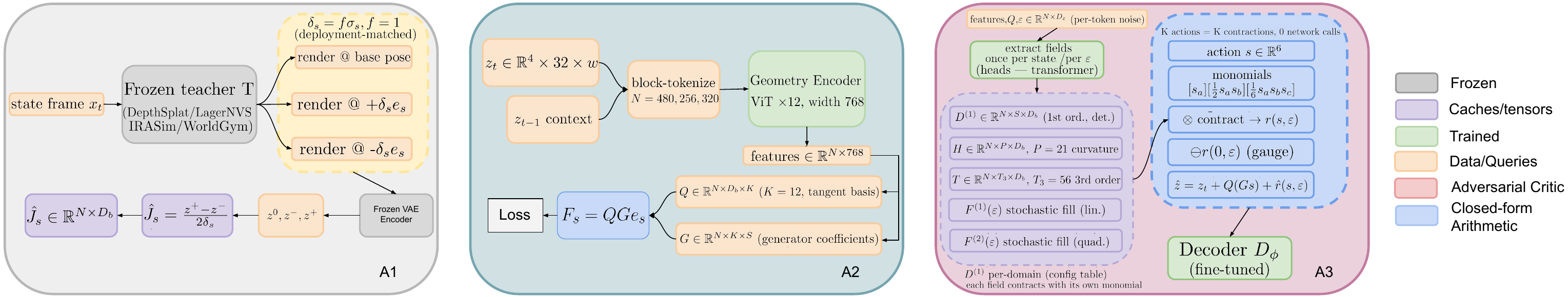}
\caption{\textbf{Architecture.} \emph{(A1)} probing; \emph{(A2)} the geometry
encoder, emitting $Q$ and $G$; \emph{(A3)} the correction head, whose coefficient
fields depend on the state and a noise draw but never on the action. Everything
in the red region is arithmetic on tensors already in memory.}
\label{fig:pipeline_a}
\end{figure}

The Method's architectural overview is presented in Figure~\ref{fig:pipeline_a}, which  draws the three parts of the method at the level of tensors. \emph{(A1)} The symmetric probe renders the generator at $\pm\delta$ along one axis and differences the two encodings, so everything the two renders share cancels, and what survives is the generator's response to the action rather than its rendering. \emph{(A2)} The geometry encoder reads the current latent and one context frame and emits an orthonormal frame $Q$ with per-axis coefficients $G$, so that all $A$ axis fields live in one rank-$m$ subspace instead of being $A$ unrelated predictions, which lets the local dimension of the manifold be estimated as an effective rank rather than assumed. \emph{(A3)} The correction head emits coefficient fields that depend on the state and a noise draw but never on the action. This property is what the cost structure of Table~\ref{tab:bridge_nfe} rests on, making it so that once the coefficients are extracted, every candidate action is a contraction against tensors already in memory, so the network runs once per state no matter how many candidates are scored, and only the winner is decoded.

\subsection{What a chart emits per-axis}
\label{app:chart_anatomy}

\begin{figure}[h]
\centering
\includegraphics[width=\textwidth]{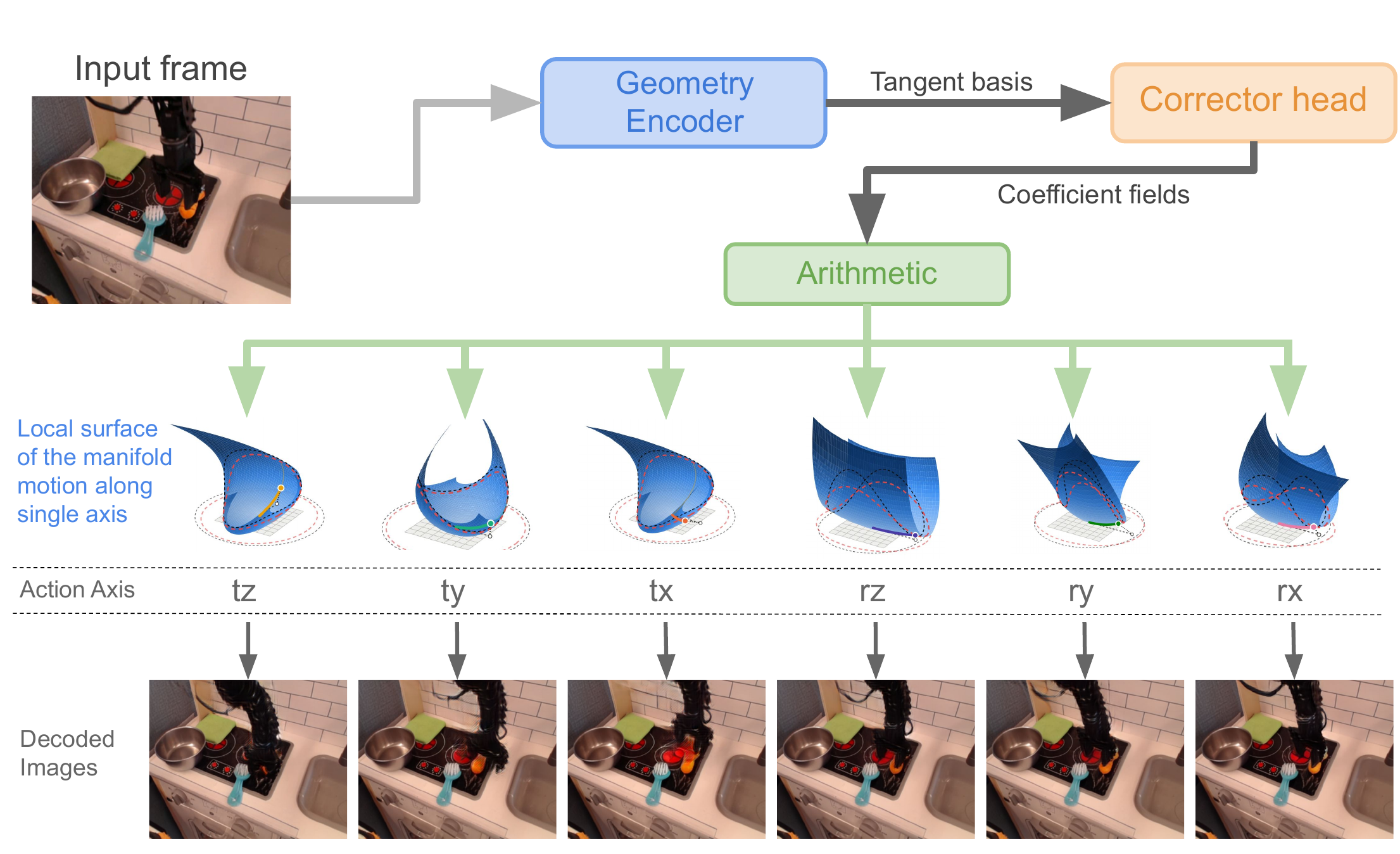}
\caption{\textbf{Extracting and tracing a local chart.} A held-out state is encoded once and the correction head's coefficient fields are extracted once. The six branches then read that one extraction along each action axis. Each branch shows a surface of the local action manifold $\mathcal{M}_x$ and a corresponding decoded frame, predicted from the step along the labeled axis. \emph{Reading a surface render:} $x$ and $y$ are commanded action in the plane spanned by the labeled axis and one companion axis, the latter present only so that the patch is a surface rather than a curve. The dashed circle is the query radius for that axis for that episode, and red dashed circle is the validity radius. The dashed line on the tangent plane is the departure from the tangent basis plane. The flat gray sheet is the first-order chart $z_t + QGa$, which is the zero-height plane in this embedding and so is flat by construction. The blue sheet adds the correction head, $z_t + QGa + \tilde r(a,\varepsilon)$. The two touch at the origin but the blue sheet leaves it with a nonzero slope. Every vertex of both sheets is arithmetic on the single extraction, no network is re-run anywhere in the figure, and only the displayed frames are decoded.}
\label{fig:axis_surfaces}
\end{figure}

Figure~\ref{fig:axis_surfaces} shows what a chart looks like once it has been extracted, on a single held-out state rather than schematically. The state enters the encoder once and the correction head emits its coefficient fields once. Then the six patches of $\mathcal{M}_x$ and the frames traced out on them are read off that single extraction.

Each panel is a two-dimensional patch, not a one-dimensional slice, and
that is a property of the drawing rather than of the measurement: a
surface needs two coordinates, so each panel sweeps the axis it is
labeled with and carries one \emph{companion} axis purely so that there
is something to plot against. 
On the visualized state from the BridgeData~V2 evaluation split released by the IRASim authors (episode 11),
the plotted pairs are (sweep axis, companion axis): $(tx,tz), (ty,tx), (tz,tx), (rx,tx), (ry,tx), (rz,tx)$. Nothing in the figure should be read as a claim about the companion axis because the swept axis is the one whose path carries the markers and whose steps are decoded into the frames next to a surface render.

Two sheets are drawn on every patch. The flat gray one is the first-order chart alone, $z_t + QGa$. It is flat because the plot's $x$ and $y$ are the two action coordinates, so the tangent chart is
exactly the zero-height plane, parallel to the floor by construction. The blue sheet adds the correction head, $z_t + QGa + \tilde r(a,\varepsilon)$, and the dashed line off the tangent plane measures how far it departs
from the tangent plane, in units of one tangent step length. The two sheets touch at $a{=}0$ but the blue one leaves with a nonzero slope, which is the correction head's linear term $L_i$.

Every vertex of both sheets is a model evaluation on a grid of commanded actions, and every one of those evaluations is a contraction against already computed coefficient fields. No part of the figure re-runs the network, the encoder pass and the extraction happen once for the whole
state, which is the same property that makes the screening cost of Table~\ref{tab:bridge_nfe} flat in the number of candidates. Only the frames are decoded, one decode each, which can be done in parallel.

The radius of App.~\ref{app:radius} is the statement that these sheets stay trustworthy out to a computable distance from the origin, and in the rendered surface the validity radius is drawn as a red dashed circle. For context, the agreement result of \S\ref{sec:evidence} is the statement that, on DL3DV, a chart learned from a different generator produces near-identical bases at the same state: mean per-axis cosine $0.80$, subspace principal cosine $0.74$, and
induced metrics matching to a relative Frobenius difference of $0.089$.

\subsection{The full instrument check on Kubric}
\label{app:kubric_extra}

For Kubric, the generator is differentiable, so
Eq.~\ref{eq:probe} can be scored against the quantity it estimates. Per-unit fields match the autodiff Jacobian at cosine $0.87$ (camera azimuth), $0.99$ (object translation) and $0.90$ (light azimuth) at $\delta{=}0.02$. Generator bias, the component of the measured field attributable to the generator rather than the scene, is $0.09$--$0.16$, and
the stochastic fill has divergence of $0.017$.

\paragraph{Disentanglement and intrinsic dimension.} The shared rank-$m$ frame of \S\ref{sec:encoder} routes different generative factors into different columns, such that off-diagonal generator overlap is
$0.19$ against $0.38$--$0.45$ measured in the trunk features the basis is read from. The one strong cross-coupling is camera azimuth against light azimuth at $-0.49$, which is expected due to high visual similarity. The effective rank of the target energy in the basis is $3.9$ of $m{=}12$ against three true action axes, and in-chart coverage is $0.94$--$0.96$.

\paragraph{Non-commutativity.} Applying two factors in both orders and comparing the endpoints gives a holonomy that orders by how much the two factors interact. Translation--light pairs score $0.53$--$0.62$ and camera rotation pairs $0.88$--$1.41$. We report the ordering rather than the magnitudes, which are in the units of the loop.

\paragraph{Render quality is not probe quality.} This is the
experiment behind our generator-selection rule. Adversarially sharpening the generator improved its reconstructions and corrupted its action derivatives, such that
probe agreement with the validated reference fell to $0.64$--$0.68$ with ${\sim}25\%$ energy inflation. The damage is not confined to the generator alone, since an Action Jet learned from the sharpened generator reproduced it faithfully, reaching per-factor held-out cosines
$0.59/0.54/0.61$ against $0.73/0.67/0.77$ for the Action Jet of its unsharpened predecessor. Perceptual losses constrain outputs, not $\partial\,\mathrm{output}/\partial\,\mathrm{action}$, so we select generators by probe agreement and never by sample quality. App.~\ref{app:agree_bridge} is the same result seen on real data, where the generator with the better LPIPS metrics (Table~\ref{tab:bridge_vis_metrics}) is the one whose chart is wrong.

\paragraph{the correction head ladder.} Table~\ref{tab:kubric_ablation}
crosses the correction-head variants with the decoder, so the two levers can be read apart. The 3-jet dominates the 2-jet on all sixteen numbers at matched decoder. Decoder fine-tuning is valuable for rendering quality, as for the 3-jet it is worth $+2.7$ to $+3.1$\,dB and $-43$ to $-61\%$ LPIPS. And the plain free-form unconstrained Transformer model, which was \emph{not} in the decoder's fine-tuning mix, renders best through it, which is the evidence that the correction head
comparison in Table~\ref{tab:kubric} is not a decoder artifact.

\subsection{The validity radius}
\label{app:radius}

\subsubsection{Definition}
\label{app:radius_def}

Along a unit direction $u$ in action space the Action Jet's prediction
displaces the latent by
\begin{equation}
  P(r) \;=\; \alpha\,r \;+\; \beta\,r^{2} \;+\; \gamma\,r^{3},
  \label{eq:cubic}
\end{equation}
with $\alpha$ collecting the tangent field $F = QG$ and the correction head's linear term, and $\beta, \gamma$ the correction head's quadratic and cubic terms. Because the correction head is a cubic in the action by construction, Eq.~\ref{eq:cubic} is exact rather than fitted, and $\alpha, \beta, \gamma$ are recovered exactly from five evaluations at $0, \pm \delta, \pm 2\delta$ for any
$\delta$. Writing $p_1 = \norm{\alpha}$, $p_2 = \norm{\beta}$, $p_3 = \norm{\gamma}$, the first-order validity radius solves $p_2r^{2} + p_3r^{3} = p_1r$:
\begin{equation}
    r_1^{*} \;=\; \frac{-p_2 + \sqrt{p_2^{2} + 4p_1 p_3}}{2p_3}.
  \label{eq:r1}
\end{equation}
It costs five evaluations of the already-extracted polynomial per noise draw, each a contraction of stored coefficient fields with no network pass and no rendering (equivalently, $\alpha, \beta, \gamma$ can be contracted directly from $F$, $L$, $H$ and $C$), so it is available per state at inference time.
It is a property of the Action Jet alone, requiring no generator and no ground-truth frame. And it is direction-dependent, such that along a demonstrated action $a$ we evaluate Eq.~\ref{eq:r1} at $u = a/\norm{a}$ rather than combining per-axis radii, because the induced metric is strongly anisotropic (App.~\ref{app:agree_nvs}, condition number ${\sim}30$). $r_1^{*}$ is invariant to a uniform rescaling of the response, meaning that scaling $\alpha, \beta, \gamma$ by a common factor leaves Eq.~\ref{eq:r1} unchanged, so a chart that simply commands less motion does not thereby report a larger radius. This matters because one of our two Bridge charts under-commands by roughly an order of
magnitude. What $r_1^{*}$ is not invariant to is the
higher-order terms collapsing on their own, 
such as $p_2, p_3 \to 0$ at fixed $p_1$, where $r_1^{*} \to \infty$ (with $p_2 = 0$, $r_1^{*} = \sqrt{p_1/p_3}$; with $p_2 > 0$ fixed, $r_1^{*} \to p_1/p_2$ and stays finite),
so a chart predicting a nearly stationary frame would report an unbounded radius while predicting nothing, which is the copy-frame failure mode. Every measurement below is guarded by the fraction of the true latent displacement the prediction actually realizes, %
$\norm{\hat z(a) - z_t}/\norm{z_{t+g} - z_t}$:
$0.71$ for the IRASim-basis chart and $0.63$ for the WorldGym-basis chart, so neither is degenerate and the
comparison in App.~\ref{app:radius_two} is between two charts that both move.

\subsubsection{Measured against real Bridge frames}
\label{app:radius_bridge}

BridgeData~V2 is a dataset of real robot arm tabletop manipulation videos, where the action from frame $i$
to frame $i{+}g$ changes with the gap $g$, and for every $g$ a real frame $i{+}g$ is available. We sweep $g = 1 \ldots 12$ over 96 states and plot everything against $|a| / r_1^{*}(a/|a|)$.

\paragraph{Coverage.} Table~\ref{tab:bridge-coverage} gives the share
of real transitions inside $r_1^{*}$. At gap~1 it is $99\%$, decaying
monotonically to $27\%$ at gap~12, and at gap~4, the horizon the correction head was trained on, it is $67\%$. Hence the radius covers 99\% of single-step Bridge transitions and two thirds of four-step ones.

\begin{table}[h]
  \centering\small
  \begin{tabular}{lcccccccc}
    \toprule
    gap & 1 & 2 & 3 & 4 & 6 & 8 & 10 & 12 \\
    \midrule
    IRASim basis   & 99.0 & 93.8 & 76.0 & 66.7 & 53.1 & 39.6 & 31.2 & 27.1 \\
    WorldGym basis & 100.0 & 90.6 & 72.9 & 60.4 & 41.7 & 26.0 & 21.9 & 22.9 \\
    \midrule
    median $|a|/r_1^{*}$ (IRASim)
                   & 0.23 & 0.42 & 0.59 & 0.76 & 0.96 & 1.16 & 1.26 & 1.37 \\
    \bottomrule
  \end{tabular}
  \caption{Real BridgeData~V2 transitions inside $r_1^{*}$ for both charts by gap. 96 states per gap. The two charts are built from different generators and use different action conventions, yet their coverage curves track each other closely.}
  \label{tab:bridge-coverage}
\end{table}

\begin{figure}[h]
  \centering
  \includegraphics[width=\linewidth]{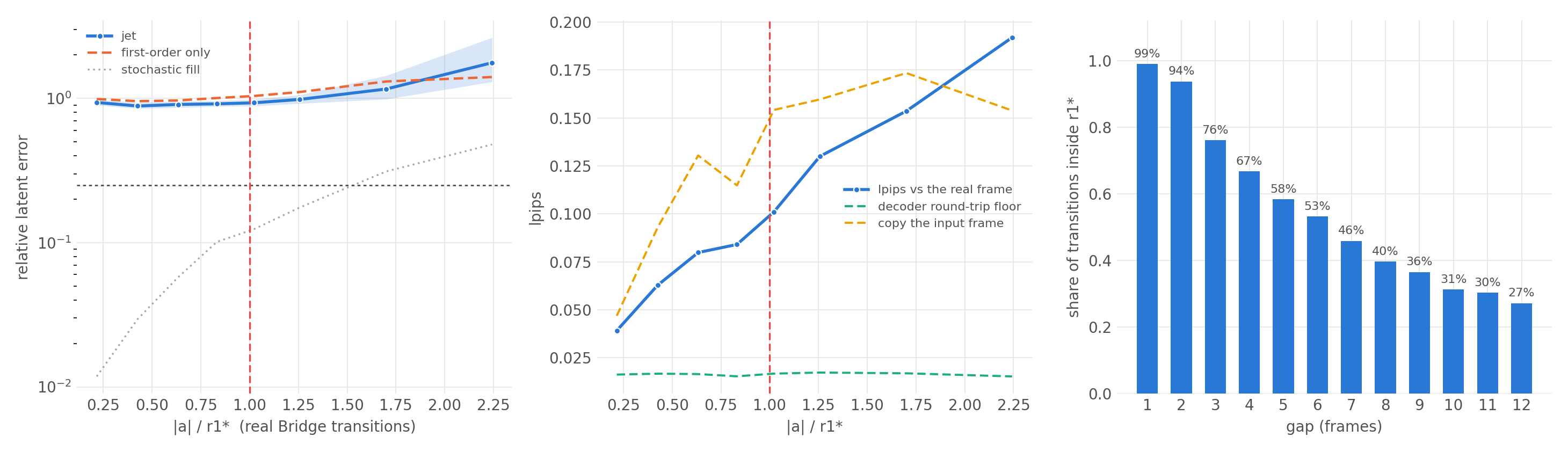}
  \caption{The radius of Action Jet (geometry encoder of which was trained on IRASim generator) measured against real frames with no generator in the loop. \emph{Left:} relative latent error of the full jet against its
  first-order truncation, with the stochastic fill level.
  \emph{Center:} LPIPS against the real frame, with the decoder round-trip floor and the copy baseline. \emph{Right:} share of real transitions inside $r_1^{*}$ by gap. Dashed vertical line is $r_1^{*}$.}
  \label{fig:bridge-curve}
\end{figure}

The operational question is where prediction stops beating the do-nothing baseline. Table~\ref{tab:bridge-curve} scores the decoded prediction against the real frame in LPIPS, alongside the decoder round-trip floor and the copy-frame baseline, which scores frame $i$ itself against frame $i{+}g$, and Figure~\ref{fig:bridge-curve} visualizes this measurement. The chart wins over copying from $|a|/r_1^{*} \approx 0.2$, peaks at $1.02$, and reaches parity at $1.89$. The decoder floor is flat at $0.016$ throughout, two to ten times below the signal. The margin is non-monotone in the step, such that at very small steps copying is nearly
right, so the advantage is small, rises to a peak near the radius, and then falls to zero. The empirical
radius is the upper end of the interval on which the jet wins, taken as the last positive margin interpolated to zero.

\begin{table}[h]
  \centering\small
  \begin{tabular}{lcccccccc}
    \toprule
    $|a|/r_1^{*}$ & 0.22 & 0.43 & 0.63 & 0.83 & 1.02 & 1.26 & 1.70 & 2.24 \\
    \midrule
    jet (LPIPS)      & 0.039 & 0.063 & 0.080 & 0.084 & 0.101 & 0.130 & 0.154 & 0.192 \\
    copy             & 0.047 & 0.093 & 0.130 & 0.115 & 0.154 & 0.160 & 0.173 & 0.154 \\
    decoder floor    & 0.016 & 0.017 & 0.016 & 0.015 & 0.017 & 0.017 & 0.017 & 0.015 \\
    \midrule
    margin           & $+0.008$ & $+0.030$ & $+0.051$ & $+0.031$
                     & $\mathbf{+0.053}$ & $+0.030$ & $+0.020$ & $-0.038$ \\
    \midrule
    \multicolumn{9}{l}{\emph{relative latent error}} \\
    full jet         & 0.937 & 0.885 & 0.908 & 0.917 & 0.932 & 0.984 & 1.162 & 1.767 \\
    first order only & 0.990 & 0.957 & 0.968 & 1.006 & 1.039 & 1.107 & 1.311 & 1.406 \\
    ratio            & 1.06 & 1.08 & 1.07 & 1.10 & 1.12 & 1.13 & $\mathbf{1.13}$ & \textit{0.80} \\
    \bottomrule
  \end{tabular}
  \caption{IRASim-basis chart on real Bridge frames, binned by $|a|/r_1^{*}$, median over 1152 transitions. Margin is copy minus jet, where positive means the chart beats predicting no motion. The margin peaks at the radius and reaches zero at $1.89\,r_1^{*}$. The lower block is an independent confirmation in latent space only.}
  \label{tab:bridge-curve}
\end{table}

\begin{figure}[t]
\centering
\includegraphics[width=0.7\textwidth]{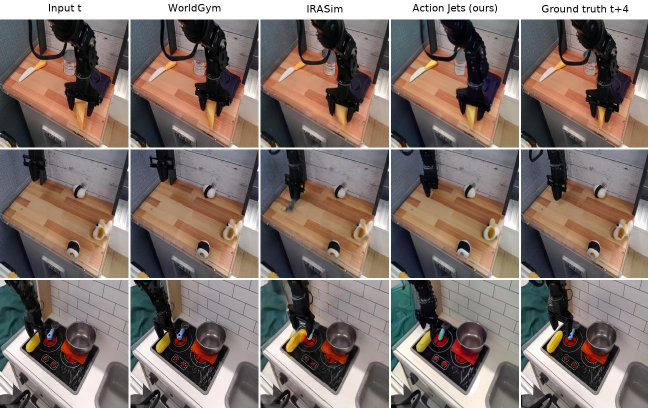}
\caption{Qualitative results and comparison between WorldGym, IRASim and Action Jet. IRASim often hallucinates or performs slightly wrong motion, while WorldGym under-commands and can show wrong movement.}
\label{fig:bridge_qual_paper_figure_small}
\end{figure}

\begin{figure}[h]
\centering
\includegraphics[width=\textwidth]{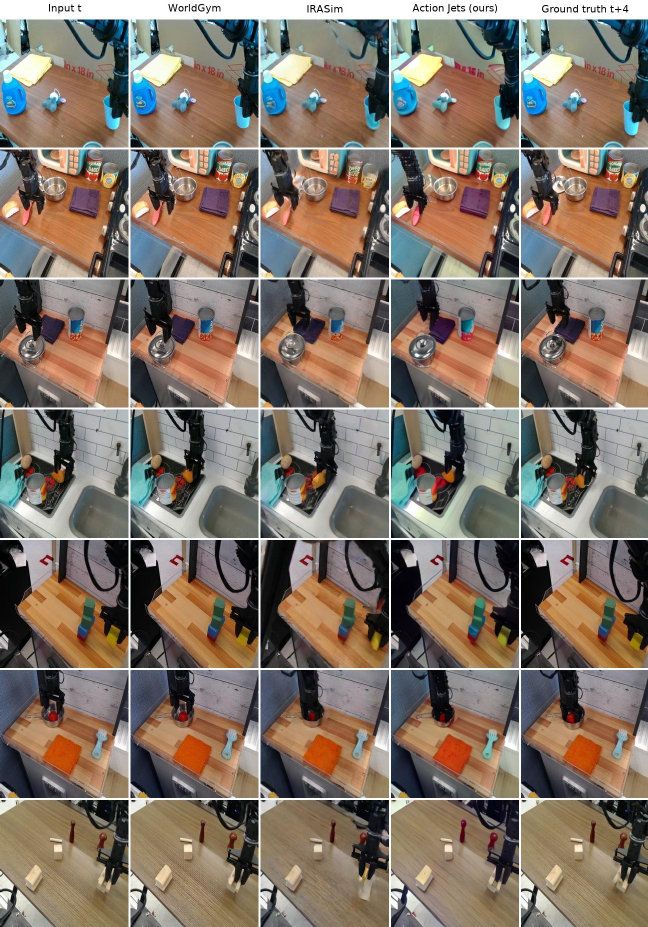}
\caption{Additional qualitative results for BridgeData~V2 open-loop rollouts. Held-out episodes under the recorded actions at gap $4$ against the real frames and the source generators. These are the rollouts scored in Table~\ref{tab:bridge_vis_metrics}.}
\label{fig:bridge_qual_sup}
\end{figure}

The last block of Table~\ref{tab:bridge-curve} shows that the ratio of the first-order-only error to the full-jet error grows from $1.06$ to $1.13$
as the step approaches and passes the radius, which means the correction does more work the further out you go, as expected, and then \emph{inverts} to $0.80$ in the last bin. Beyond about $2\,r_1^{*}$ the cubic term is not merely uninformative, but extrapolating it is worse than truncating the expansion. Figure~\ref{fig:bridge_qual_paper_figure_small} shows
what horizon $4$ predictions look like compared to IRASim and WorldGym baselines. We can see that IRASim sometimes moves incorrectly or hallucinates, while WorldGym under-commands and is almost static. Additional qualitative results can be found in Figure \ref{fig:bridge_qual_sup}.

\subsubsection{The radius is a property of the chart, not of the generator}
\label{app:radius_two}

\begin{figure}[h]
  \centering
  \includegraphics[width=\linewidth]{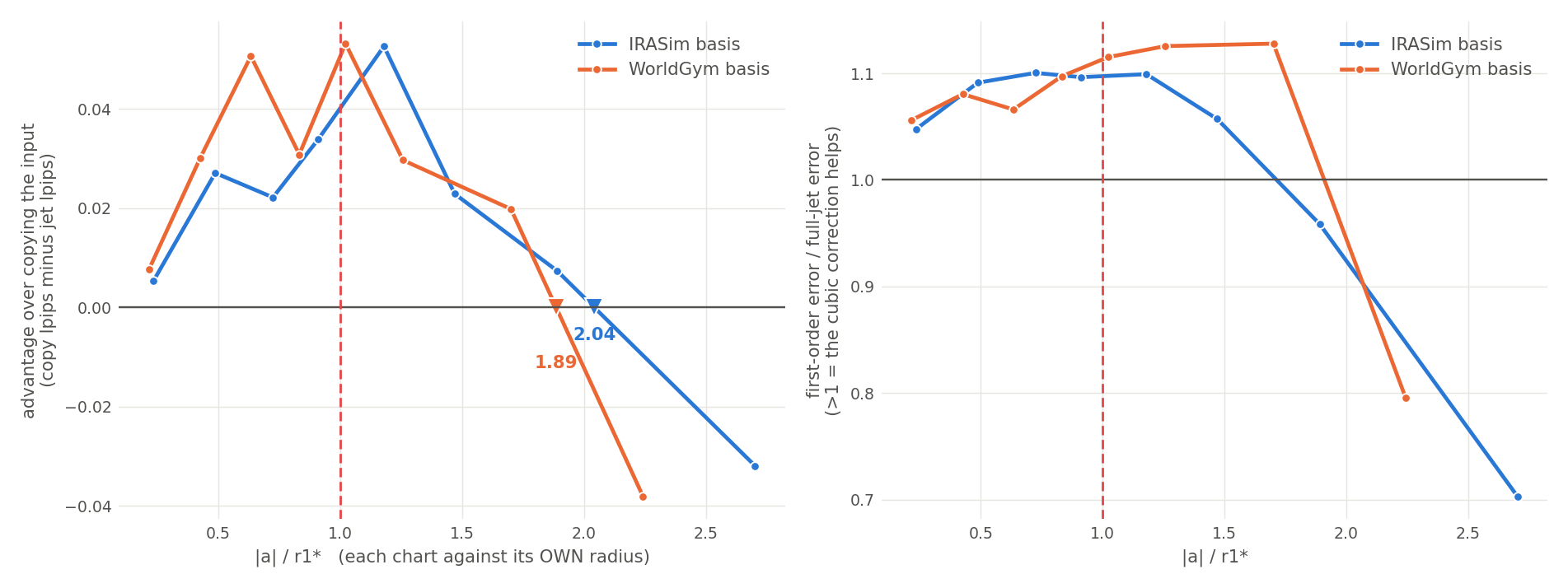}
  \caption{Advantage over the copy baseline for two Bridge charts built
  from different generators, each against its own $r_1^{*}$. The charts
  have nearly orthogonal tangent fields and translation gains differing
  by $10\times$, yet both stop beating the baseline at about twice
  their own predicted radius.}
  \label{fig:two-charts}
\end{figure}

\begin{figure}[h]
  \centering
  \includegraphics[width=\linewidth]{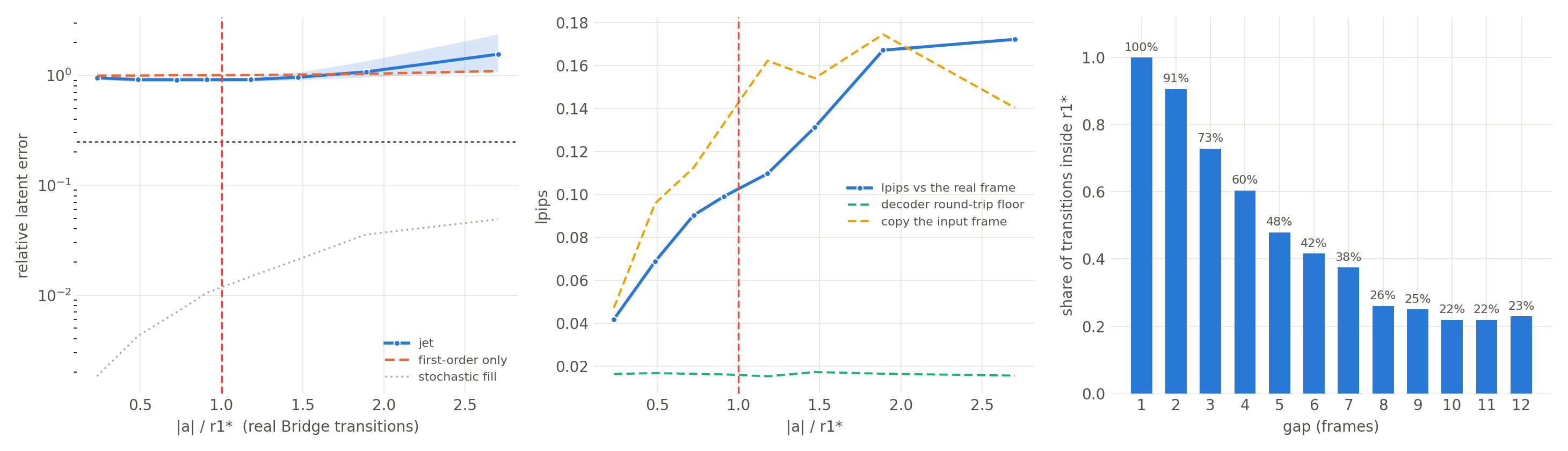}
  \caption{The WorldGym-basis chart, same three panels as
  Fig.~\ref{fig:bridge-curve}: latent error of the full jet against its first-order truncation, LPIPS against the real frame with the decoder floor and the copy baseline, and coverage inside $r_1^{*}$ by gap. This is the negative control of App.~\ref{app:agree_bridge}, a
  chart that anti-selects downstream and does not anchor on real transitions, and its radius is nonetheless calibrated. Being well calibrated about your own limits and charting the right object are separate properties, which is why we measure both.}
  \label{fig:bridge-wg}
\end{figure}

We repeat the whole validity radius measurement on the
second Bridge chart learned from WorldGym probes. The two charts, one learned from IRASim and the other from WorldGym, are about as unlike as two charts of one dataset can be, as they have nearly orthogonal tangent fields
(App.~\ref{app:agree_bridge}, mean per-axis cosine $0.04$), translation gains a factor of ten apart, stochastic fill differing by an order of magnitude. And the WorldGym one is the chart that fails to anchor on real transitions at all. If $r_1^{*}$ were an artifact of one generator's idiosyncrasies, the two would not break down at comparable multiples of their own radii, but they do. Both charts cross the copy baseline at $1.89$ and $2.04$ times their own $r_1^{*}$, both peak at $+0.053$ LPIPS of advantage, at $1.02$ and $1.18$, and both show the correction head's advantage growing to about $1.1\times$ and then inverting in the last bin. Both charts give a positive per-state rank correlation, $+0.61$ and $+0.55$, and their empirical-to-predicted ratios, $1.72$ and $2.02$, agree to within $18\%$. Table~\ref{tab:two-charts} in the main paper collects these, Fig.~\ref{fig:two-charts} draws both margin curves on the shared normalized abscissa, and Fig.~\ref{fig:bridge-wg} is the WorldGym run in the same three panels as Fig.~\ref{fig:bridge-curve}, including the right-hand panel that shows the share of real transitions inside $r_1^{*}$ at each gap, which is the coverage claim of \S\ref{sec:radius_emp} drawn per chart.

\subsubsection{Generator probes on DL3DV}
\label{app:radius_dl3dv}

\begin{figure}[h]
  \centering
  \includegraphics[width=\linewidth]{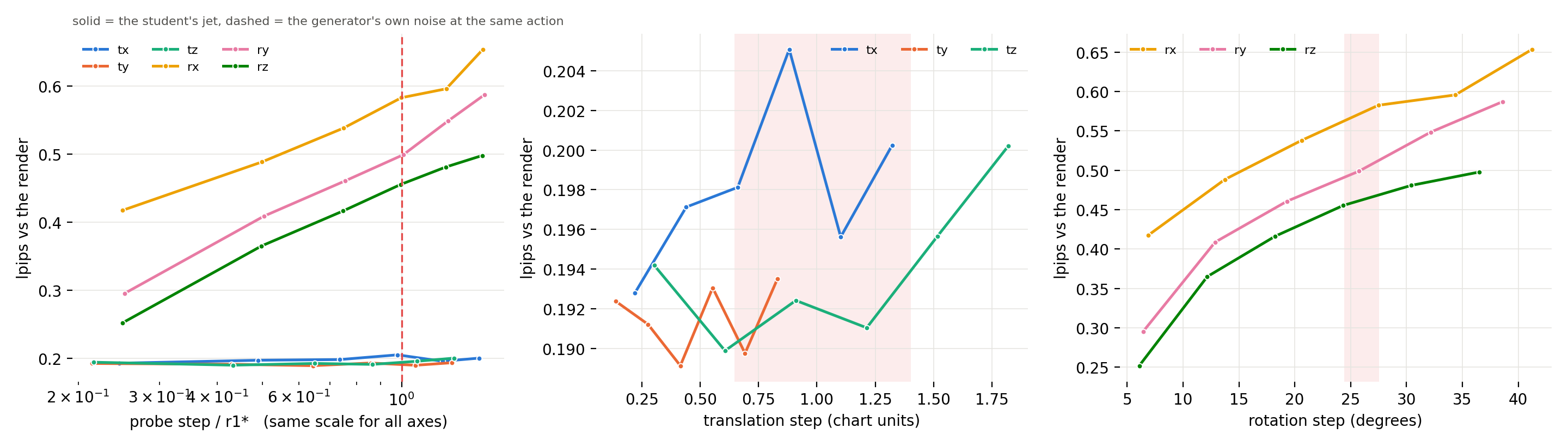}
  \caption{DL3DV degradation across the probe ladder. The chart-to-generator gap is what the ratio of $0.74$--$0.85$ quoted above measures, and it is stable across the whole ladder. \emph{Left:} all six axes on the
  shared normalized abscissa, log scale, with $r_1^{*}$ marked, with the flat translation lines measured in Table~\ref{tab:dl3dv-probe-motion}. \emph{Center and right:} the same data against the physical step, split because $0.5$ chart units and $0.5^{\circ}$ are not the same picture.}
  \label{fig:degradation}
\end{figure}

In BridgeData~V2 the whole arm is not always visible, and one of the generators (WorldGym) is a negative control. Hence we use the DL3DV dataset of camera motion in static indoor and outdoor scenes to measure generator agreement and learned chart consistency. The camera translations and rotations are very different in terms of the visual change that they induce per unit change in action, so to conduct our measurements we instead step each axis to a fraction of its own $r_1^{*}$, rendering six intervals at $\{0.25 \ldots 1.5\}\,r_1^{*}$, and in physical units the rotation intervals run $6.9^{\circ}$ to $41.2^{\circ}$ and the translations $0.14$ to $1.82$ chart units. 
DL3DV translation probes change the image less than the decoder's own round-trip floor, even at $1.5\,r_1^{*}$, 
while rotation probes move it $1.5$ to $7$ times above that floor (Table~\ref{tab:dl3dv-probe-motion}).
This is a property of the
benchmark since DL3DV scenes are distant enough that camera translation produces little parallax, and it is why the Bridge measurement is the primary one for the validity radius. We report DL3DV degradation on the
rotation axes, Fig.~\ref{fig:degradation} also draws the flat translation lines for reference.
On the rotation axes the Action Jet stays below the induced motion at every rung, beating the copy-frame control, at a ratio of predicted error to induced motion of $0.74$--$0.85$ that is stable across a $6\times$ range of step size. The cubic correction reduces the latent error by $1.9$--$3.8\times$ there, far more than on Bridge, because the rotation response is much more nonlinear.

\begin{table}[h]
  \centering\small
  \begin{tabular}{lccc}
    \toprule
    probe axes & probe motion (LPIPS) & decoder floor & verdict \\
    \midrule
    translations $t_x, t_y, t_z$ & 0.015--0.109 & 0.12--0.15
      & below the floor \\
    rotations $r_x, r_y, r_z$    & 0.22--0.85   & 0.12--0.15
      & $1.5$--$7\times$ above \\
    \midrule
    \emph{Bridge, for contrast} & \multicolumn{2}{c}{floor $0.016$;
      every axis moves the image} & measurable \\
    \bottomrule
  \end{tabular}
  \caption{Did the DL3DV probe move the picture? LPIPS between each probe render and the generator's own render at the base action, against the decoder round-trip floor on the same frames. 
  A decoded measurement below its own instrument's noise floor carries no information, which is what rules the translation axes out of the pixel-space radius measurement on this benchmark (Fig.~\ref{fig:degradation}). The agreement results of Tables~\ref{tab:agreement} and~\ref{tab:agreement-nvs-full} compare latent fields directly and never pass through the decoder, so this floor does not limit them. Range over the six ladder intervals, $\{0.25 \ldots 1.5\}\,r_1^{*}$, $n = 64$ states.}
  \label{tab:dl3dv-probe-motion}
\end{table}

Fitting Eq.~\ref{eq:cubic} to the generator's rendered probe responses and evaluating Eq.~\ref{eq:r1} on that fit gives a radius that involves no threshold and no chart, a genuinely independent estimate of the same quantity. Across all six axes the generator-implied radius is
$0.48 \pm 0.03$ of the chart's (Table~\ref{tab:dl3dv-calib}), a strikingly constant factor. The per-state rank correlation is positive on the translation axes ($+0.24$ to $+0.40$) and absent or negative on the rotation axes ($-0.27$ to $-0.02$). This is consistent with the rotation fits being taken across a range where the response is already strongly nonlinear, since almost all rotations in DL3DV produce strong parallax and disocclusions. We report the DL3DV calibration as a scale result and rest the per-state claim on Bridge, where it is measured against real frames.
The last column of Table~\ref{tab:dl3dv-calib} shows that on the DL3DV rotation axes $r_1^{*}$ sits $7$--$27\times$ beyond the 95th-percentile rotation the correction head saw in training, so those radii are extrapolations of Eq.~\ref{eq:r1} and the ladder exists precisely to put rendered evidence underneath that range, which is done by the $1.25$ and $1.5\,r_1^{*}$ intervals. On Bridge no extrapolation is involved because the ramp reaches $3.1\times$ the radius at $p_{95}$ with a real frame at every interval.

\begin{table}[h]
  \centering\small
  \begin{tabular}{lccccc}
    \toprule
    axis & $r_1^{*}$ (chart) & generator-implied & ratio
    & $\rho$ per state & $r_1^{*}/p_{95}(|a|)$ \\
    \midrule
    $t_x$ & 0.900 & 0.446 & 0.50 & $+0.40$ & 0.7 \\
    $t_y$ & 0.646 & 0.283 & 0.44 & $+0.24$ & 1.5 \\
    $t_z$ & 1.403 & 0.617 & 0.44 & $+0.28$ & 1.7 \\
    $r_x$ & 0.480 & 0.250 & 0.52 & $-0.02$ & 18.4 \\
    $r_y$ & 0.446 & 0.220 & 0.49 & $-0.27$ & 6.8 \\
    $r_z$ & 0.426 & 0.217 & 0.51 & $-0.20$ & 26.7 \\
    \midrule
    mean  &       &       & \textbf{0.48} & & \\
    \bottomrule
  \end{tabular}
  \caption{DL3DV: the learned chart's $r_1^{*}$ against the radius implied by the generator's own probe responses, per axis, $n = 64$ states. 
  The last column gives $r_1^{*}$ as a multiple of the 95th percentile of the recorded action magnitude. Radii are for the DepthSplat chart, mean over the 64 ladder states, so they differ from Table~\ref{tab:radius}, which reports medians.}
  \label{tab:dl3dv-calib}
\end{table}

Table~\ref{tab:radius} gives $r_1^{*}$ and the stochastic fill floor per axis for Action Jets, in normalized chart units. The two DL3DV charts, charted from unrelated generators, agree to within about $10\%$ on every axis, showing the radius consequence of the induced-metric agreement in App.~\ref{app:agree_nvs}. 
And the fill floors describe the data where camera rotations in DL3DV bring new content into the frame, while end-effector motion in BridgeData~V2 moves a gripper across a static background, and the fill floor tracks that disocclusion, $3.5$--$8.7$ per unit radius on camera rotations against at most $0.97$ anywhere on the robot motion.

\subsection{Learned chart analysis}
\label{app:agree}

\paragraph{A protocol that makes fields comparable.}\label{app:protocol} A field becomes evidence about $\mathcal{M}_x$ only when fields from different generators can be compared entry by entry, so we fix everything except the generator: the states, the image window (on scenes, the largest central square that every generator renders, to within $0.2\%$), the probe displacement (App.~\ref{app:controls_guide}), the latent space, and the action, mapped into each generator's convention by a kinematic change of coordinates certified before use (median cosine $+1.0000$ over $2{,}405$ transitions, against $-0.72$ for the identity map). We score the \emph{agreement} of two generators as the cosine similarity between their fields for the same state and axis, computed over all tokens and summarized by the median over states and the mean over axes; it ignores magnitude on purpose, and random fields score $0.001$. For stochastic generators there is also a ceiling: with split-half reliabilities $\rho_A, \rho_B$, a true agreement $c$ is measured as about $c\sqrt{\rho_A \rho_B}$, and we report that ceiling beside the raw value. App.~Table~\ref{tab:controls} lists eleven controls and the artifact each rules out.

\subsubsection{Generators' agreement noise}
\label{app:agree_noise}

Two generators cannot agree beyond their own reproducibility, hence we measure it directly by rendering the probes twice with different seeds and comparing each generator with itself. Rendering the LagerNVS ladder
a second time at a different seed reproduces the first to seven significant figures, per-axis 
$\cos(\widehat J^{(0)}, \widehat J^{(1)})
= 0.99999952$--$0.99999992$, target norms
agreeing to $10^{-6}$, because the LagerNVS backbone is feed-forward and deterministic. DepthSplat is deterministic by construction. Both DL3DV generators have reliability $\rho = 1.000$, so the observed generator--generator cosine is the true signal agreement with no attenuation, and the Action Jets' agreement cannot be inflated by noise canceling. This is a stronger position than a measured noise floor would have given us.

On Bridge, WorldGym is stochastic and we measure its reliability from a seed replicate, where $\rho_{\mathrm{WG}} = 0.79$--$0.89$ per axis. IRASim is a video diffusion model too slow to re-render at the required scale, so we bound it instead, and the bound is what lets the Bridge negative stand without a replicate. We model a measured target as signal plus independent noise, and a learned chart that recovers the signal exactly scores $\cos(F, \widehat J) = \sqrt{\rho}$, and any worse learned chart scores less, so
\begin{equation}
  \rho \;\ge\; \cos(F, \widehat J)^{2},
  \label{eq:rho_bound}
\end{equation}
a lower bound on a generator's reproducibility computed from the learned chart's fidelity. From the measured $\cos(F_{\mathrm{IRASim}}, \widehat J_{\mathrm{IRASim}})$ this gives
$\rho_{\mathrm{IRASim}} \ge 0.08$--$0.51$ per axis. The bound is loose by construction, so it can only make the
ceiling in App.~\ref{app:agree_bridge} smaller and therefore the negative result harder to claim.

\subsubsection{Chart agreement on DL3DV}
\label{app:agree_nvs}

Table~\ref{tab:agreement} in the main paper gives the medians and Table~\ref{tab:agreement-nvs-full} adds the interquartile ranges. Against a state-permutation null at $0.00 \pm 0.015$ on every axis, two charts trained on DepthSplat and LagerNVS probes agree at $0.77$--$0.84$ per
axis, mean $0.80$, while their deterministic generators agree at mean $0.75$. On four
of six axes the learned charts agree with each other more than their generators DepthSplat and LagerNVS do.

\begin{table}[h]
  \centering\small
  \begin{tabular}{lccc}
    \toprule
    axis & Action Jets $\cos(F_A, F_B)$ & generators $\cos(\widehat J_A, \widehat J_B)$
    & null \\
    \midrule
    $t_x$ & 0.77 [0.71, 0.80] & 0.72 [0.62, 0.79] & \phantom{-}0.00 \\
    $t_y$ & 0.81 [0.74, 0.85] & 0.73 [0.63, 0.79] & $0.00$ \\
    $t_z$ & 0.77 [0.70, 0.81] & 0.68 [0.56, 0.75] & $0.00$ \\
    $r_x$ & 0.84 [0.78, 0.87] & 0.69 [0.64, 0.73] & $0.00$ \\
    $r_y$ & 0.82 [0.79, 0.85] & 0.84 [0.80, 0.87] & \phantom{-}0.00 \\
    $r_z$ & 0.79 [0.72, 0.84] & 0.86 [0.82, 0.89] & $0.00$ \\
    \midrule
    mean  & \textbf{0.80} & 0.75 & 0.00 \\
    \bottomrule
  \end{tabular}
  \caption{Cross-generator tangent agreement on DL3DV, median [IQR] over 256 states, probed at $0.25\,r_1^{*}$ per axis. Action Jets column compares the two Action Jet charts learned from DepthSplat and LagerNVS probes. Generators column compares the DepthSplat and LagerNVS generators that supervised them, both of which are deterministic (App.~\ref{app:agree_noise}).}
  \label{tab:agreement-nvs-full}
\end{table}

Principal cosines between the two six-dimensional tangent 
spaces are 
$(0.87, 0.84, 0.81, 0.75, 0.68, 0.49)$, mean $0.74$, so the subspaces largely coincide. The induced metrics $\Gamma_{ij} = \langle F_i, F_j \rangle$ agree to a relative Frobenius difference of $0.089$ (IQR $[0.066, 0.115]$) with identical eigenvalue ordering, and both charts are similarly anisotropic (condition number $29.1$ against $32.0$). Since $\Gamma$ sets the first-order coefficient in every direction, this sits directly upstream of App.~\ref{app:radius}, showing that the two charts do not merely point the same way, but also assign the same lengths. It is also why $r_1^{*}$ is evaluated along the traveled direction rather than combined from per-axis radii, since with a condition number near $30$, an isotropic summary of the radius would be wrong by a factor of five.

Two charts supervised by different generators being closer to each other than the generators are is consistent with each encoder discarding the component of its supervision that the other generator does not share, since the
generator-specific part of a measured field is, from the encoder's point of view, unpredictable structure, and a fit under the loss of Eq.~\ref{eq:enc} has no reason to preserve it. This is not explained by shared architecture alone, because the two charts use the same encoder architecture on Bridge as well, and those charts do not agree (App.~\ref{app:agree_bridge}), making Bridge's negative result function as the control for this positive one.
The measurement above steps each axis to $0.25$ of its own radius. The same comparison at a $1\sigma$ probe on a band-square window and 148 states (Table~\ref{tab:agreement_sigma}) gives generators $0.84$ and Action Jets $0.82$, arriving at the same conclusion, with the Action Jets' margin over the generators reversed in sign and within the spread. The two measurements are not interchangeable and we do not average them, since per
App.~\ref{app:radius_dl3dv}, $1\sigma$ is a different physical step on every axis and sits past breakdown on the translations. We report the $r_1^{*}$-normalized ladder in the main paper because it is the only scale at which all six axes are comparably far from their own breakdown, and quote the $\sigma$-probe here as the robustness check.

\begin{table}[h]
\centering\footnotesize
\begin{tabular}{lcccccc|cc}
\toprule
pair & $t_x$ & $t_y$ & $t_z$ & $r_x$ & $r_y$ & $r_z$ & mean & $n$ \\
\midrule
\multicolumn{9}{l}{\emph{generator $\leftrightarrow$ generator}}\\
DL3DV: DepthSplat $\leftrightarrow$ LagerNVS & .81 & .82 & .81 & .88 & .89 & .85 & \textbf{.84} & 148 \\
Bridge: IRASim $\leftrightarrow$ WorldGym (span) & -- & -- & -- & -- & -- & -- & \textbf{.036} & 148 \\
\midrule
\multicolumn{9}{l}{\emph{generator $\leftrightarrow$ probed baseline}}\\
DL3DV: LagerNVS $\leftrightarrow$ FrameCrafter & .67 & .60 & .67 & .78 & .79 & .69 & .70 & 148 \\
DL3DV: DepthSplat $\leftrightarrow$ FrameCrafter & .59 & .55 & .60 & .72 & .75 & .63 & .64 & 148 \\
\midrule
\multicolumn{9}{l}{\emph{Action Jet $\leftrightarrow$ Action Jet}}\\
DL3DV pair, charted separately & .79 & .83 & .78 & .86 & .84 & .82 & \textbf{.82} & 148 \\
Bridge pair, charted separately & $-$.02 & .05 & .07 & $-$.10 & .09 & .14 & \textbf{.039} & 148 \\
\bottomrule
\end{tabular}
\caption{Robustness check for charts agreement: the same comparisons at a $1\sigma$ probe, band-square window on scenes, 148 states. FrameCrafter is probed but not charted. Conclusions are unchanged, the absolute values are not comparable with Table~\ref{tab:agreement-nvs-full}, because the two ladders step the axes by different physical amounts.}
\label{tab:agreement_sigma}
\end{table}

\subsubsection{Chart disagreement on BridgeDataV2}
\label{app:agree_bridge}

Listed are the measurements \S\ref{sec:bridge} refers to regarding the generator and Action Jet chart disagreement on BridgeData~V2. We learn a chart from IRASim and another from WorldGym, then probe both on 148 identical held-out Bridge states, map each generator's action convention into a common frame, and compare the tangent fields exactly as on DL3DV. The two Action Jet charts agree at $-0.09$ to $0.12$ per
axis at the permutation null, and so do their \emph{generators}, at $-0.04$ to $0.05$. As established by Eq.~\ref{eq:rho_bound}, this result is not a noise artifact. Two generators measuring a common signal would agree at
$\sqrt{\rho_A \rho_B} \cdot \cos(s_A, s_B)$. With $\rho_{\mathrm{WG}}$ measured and $\rho_{\mathrm{IRASim}}$ bounded below, the ceiling is at least $0.25$--$0.67$ depending on axis, against an observed $-0.04$ to $0.05$ (Table~\ref{tab:bridge-agree}). The noise-free tangent
responses of the two generators therefore have cosine at most
$0.04$--$0.21$, so IRASim and WorldGym genuinely define different tangent structures on Bridge. Each Action Jet chart is meanwhile faithful to its own
generator, with $\cos(F, \widehat J) = 0.28$--$0.71$ for IRASim and
$0.49$--$0.65$ for WorldGym, the latter against a measured ceiling of $\sqrt{0.79} = 0.89$, so the disagreement is inherited rather than introduced.

\begin{table}[h]
  \centering\small
  \begin{tabular}{lcccccc}
    \toprule
    axis & Action Jets & generators & $\rho_{\mathrm{IRASim}} \ge$
    & $\rho_{\mathrm{WG}}$ & ceiling $\ge$ & $\cos(s_A, s_B) \le$ \\
    \midrule
    $t_x$ & $-0.02$ & $-0.02$ & 0.48 & 0.87 & 0.65 & 0.04 \\
    $t_y$ & \phantom{-}0.06 & \phantom{-}0.03 & 0.51 & 0.87 & 0.67 & 0.05 \\
    $t_z$ & \phantom{-}0.08 & \phantom{-}0.05 & 0.40 & 0.89 & 0.59 & 0.08 \\
    $r_x$ & $-0.09$ & $-0.04$ & 0.19 & 0.81 & 0.39 & 0.11 \\
    $r_y$ & \phantom{-}0.09 & \phantom{-}0.04 & 0.23 & 0.84 & 0.44 & 0.10 \\
    $r_z$ & \phantom{-}0.12 & \phantom{-}0.05 & 0.08 & 0.79 & 0.25 & 0.21 \\
    \bottomrule
  \end{tabular}
  \caption{The two generators for BridgeData~V2 do not share a tangent signal. $\rho_{\mathrm{WG}}$ is measured from a seed replicate, while $\rho_{\mathrm{IRASim}}$ is the lower bound $\cos(F,\widehat J)^{2}$ of Eq.~\ref{eq:rho_bound}, since IRASim is a diffusion model too slow to re-render. The final column is the implied upper bound on the cosine between the two generators' noise-free responses.}
  \label{tab:bridge-agree}
\end{table}

Because the two generators express actions in different conventions, a per-axis cosine could in principle understate their agreement, since a consistent remixing of the six axes would destroy it while leaving the underlying surface identical. We therefore also compare the subspaces their six fields span, with a principal-angle overlap
invariant to any such remixing ($1$ for identical subspaces). It is $0.036$, against $0.573$ when one generator is probed twice and $0.001$ for random fields, and the disagreement passes unchanged to the
Action Jets ($0.039$). A per-state $6{\times}6$ mixing fit gives the same result ($0.12$ recovered, against $0.78$ on the DL3DV pair).

Table~\ref{tab:bridge_contrast} shows that the Action Jet chart extracted from IRASim is the one that belongs to the data. Fit to the generator cannot tell the two
charts apart ($0.513$ and $0.542$), so any distillation-style
metric would call them equally good. Anchoring against real held-out transitions separates them immediately, showing that chart learned from IRASim scores
cosine $+0.249$ against a rolled-action control of $-0.014$, a gap of $+0.263$, with least-squares gain $0.74$, while the one learned from WorldGym scores $-0.011$ against a control of $+0.088$, a gap of $-0.099$, with gain $-0.10$, showing it is a chart whose best linear fit to the data points backwards. Re-probing raised the WorldGym Action Jet's fit to its own generator from $0.39$ to $0.54$ and did not move its anchoring, so fitting the generator better did not bring it closer to the data. We keep this result as a negative control for everything downstream, and as the dissimilar chart of App.~\ref{app:radius_two}.

\begin{table}[h]
\centering\footnotesize
\begin{tabular}{lcc}
\toprule
 & IRASim Action Jet & WorldGym Action Jet \\
\midrule
held-out fit to own generator's fields & $0.513$ & $0.542$ \\
real-transition anchoring (cos, mid-band, $n{=}48$) & $\mathbf{+0.249}$ & $\mathbf{-0.011}$ \\
rolled-action control & $-0.014$ & $+0.088$ \\
anchoring gap (true $-$ control) & $\mathbf{+0.263}$ & $\mathbf{-0.099}$ \\
least-squares gain $\kappa^{*}$ (median) & $0.74$ & $-0.10$ \\
tangent vs.\ copy floor (PSNR, dB) & $13.49$ vs.\ $12.94$ & $12.84$ vs.\ $12.94$ \\
\bottomrule
\end{tabular}
\caption{Fit to the generator cannot tell the two Action Jets apart, so a distillation metric would call them equally good, but anchoring against real held-out transitions separates them at once. The rolled-action control pairs each state's fields with another state's action, so it measures the anchoring any two arrays would show. The least-squares gain $\kappa^{*}$ is the scalar that best maps the
chart's predicted displacement onto the real one, where negative means the chart's best linear fit to the data points backwards.}
\label{tab:bridge_contrast}
\end{table}

\subsection{Extended DL3DV results}
\label{app:nvs_extra}

Table~\ref{tab:nvs_all} is the full per-gap grid behind
Table~\ref{tab:nvs_visual_metrics}, with SSIM, tail LPIPS over the hardest third of states, and paired bootstrap intervals against our LagerNVS-probed learned chart. Table~\ref{tab:nvs_abl} ablates the chart
itself at all four gaps. The charts learned from DepthSplat and LagerNVS differ by at most $0.006$ LPIPS and
$0.09$\,dB at every gap, which is consistent with the agreement result of App.~\ref{app:agree_nvs}, showing that two charts that share no generator make the same predictions.

The rows showing results for autoregressively composed prediction lose $0.045$ to $0.103$ LPIPS against a
single extrapolated query on every interval, because
re-encoding each decoded frame compounds error faster than the calibrated jet loses accuracy in extrapolation. 
Bridge behaves the other way, where composed rollouts edge out single extraction (App.~\ref{app:budget}), and the radius explains the split: on Bridge a single step is almost always inside $r_1^{*}$ ($99\%$) while a four-step chunk often is not ($67\%$, Table~\ref{tab:bridge-coverage}), so composing keeps every query inside, while on DL3DV even one step can exceed the translation radii ($r_1^{*}/p_{95} = 0.7$--$1.7$, Table~\ref{tab:dl3dv-calib}), so there is no inside to stay in and the extra encode is pure loss.

\begin{figure}[h]
\centering
\includegraphics[width=\textwidth]{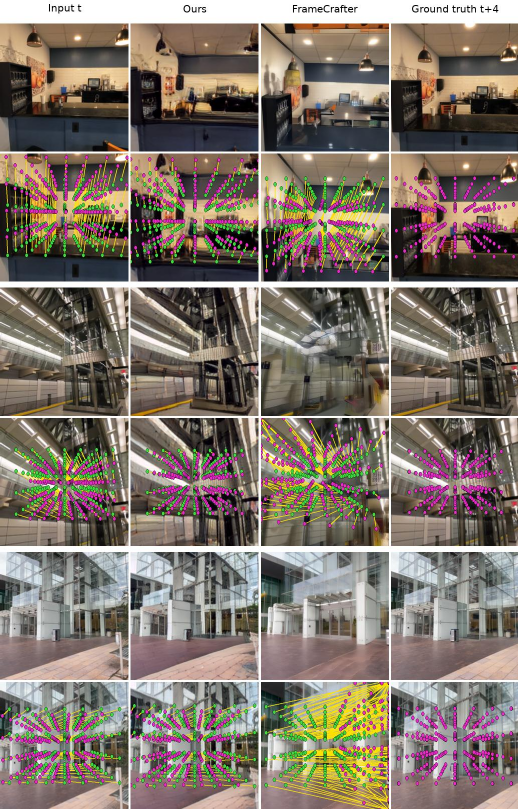}
\caption{DL3DV forward prediction and geometric consistency qualitative results. Held-out scenes from DL3DV Benchmark dataset at target frame $t+4$ compared to Action Jet and FrameCrafter predictions. The overlay marks the MASt3R correspondences measured in Table~\ref{tab:nvs_reprojection_metrics}. Action Jet may blur high-frequency details or leave slight artifacts at repeated patterns, while the diffusion baseline shows camera-motion and reprojection errors.}
\label{fig:nvs_qual_sup1}
\end{figure}

\begin{figure}[h]
\centering
\includegraphics[width=\textwidth]{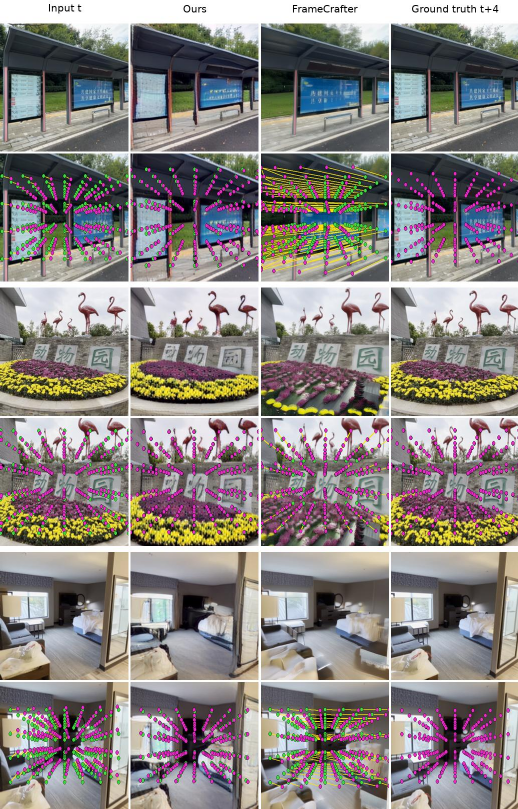}
\caption{Additional DL3DV forward prediction and geometric consistency qualitative results, continued from Fig.~\ref{fig:nvs_qual_sup1}.}
\label{fig:nvs_qual_sup2}
\end{figure}

Table~\ref{tab:nvs_reprojection_metrics}
scores geometry rather than pixels, and we describe the full evaluation protocol here. Each predicted frame is paired with the real frame at the same target pose and the pair is matched with MASt3R. Then PnP on those matches recovers the relative camera between prediction and ground truth frame, whose ground truth error is the identity, and we report the median rotation error and the virtual-correspondence reprojection error (VCRE) of that camera. A frame matched against itself gives the instrument's own
noise at $0.12^{\circ}$ and $0.4$\,px, below which no method can be distinguished. And the copy control scores the input frame as if it were the prediction, which is exactly the camera motion a predict-nothing model leaves uncorrected. Without it, a low rotation error would be unreadable, since a method that changes nothing on a slow camera path would score well. All images use the common 256-pixel window. Figures~\ref{fig:nvs_qual_sup1} and~\ref{fig:nvs_qual_sup2} show the predictions these numbers summarize, with the MASt3R correspondences that Table~\ref{tab:nvs_reprojection_metrics} scores drawn on top.

\subsection{Instrument ladders, controls, and encoder fidelity}
\label{app:controls_guide}

\label{sec:scope}
Table~\ref{tab:ladder_full} probes DepthSplat at six displacements from $0.25\sigma$ to $6\sigma$ and compares per-unit fields at adjacent intervals. Over this $24\times$ range the surface never reads as flat: the mean adjacent-rung cosine is $0.69, 0.59, 0.41, 0.50, 0.29$, and the
most curved axis is $r_y$ (pan), the one with the largest image motion per unit action. Pan $r_y$ is also the one axis whose cosine is already low at the smallest interval ($0.33$ at $0.25\!\leftrightarrow\!0.5$ against $0.67$--$0.83$ for the others), so on that axis there is no interval in this
ladder that behaves linearly at all. The one non-monotone entry $2\!\leftrightarrow\!3$ reads higher than
$1\!\leftrightarrow\!2$ because the intervals are not uniformly spaced. An Action Jet trained at
$\delta{=}1\sigma$ scores $0.82$ against held-out fields probed at $1\sigma$ and falls away in both directions, to $0.58$ one octave below and $0.43$ one octave above, reaching $0.11$ at $6\sigma$. A chart extracted at one displacement is a chart of that displacement. Hence we never compare generators probed at different displacements, because two
different generators probed at the same $\delta$ agree more
($0.84$, Table~\ref{tab:agreement_sigma}) than one generator agrees with itself across neighboring scales ($0.41$).

\begin{table}[h]
\centering\small
\begin{tabular}{lcccccc|c}
\toprule
rung pair & $t_x$ & $t_y$ & $t_z$ & $r_x$ & $r_y$ & $r_z$ & mean \\
\midrule
$0.25\!\leftrightarrow\!0.5$ & .67 & .78 & .73 & .83 & .33 & .83 & .69 \\
$0.5\!\leftrightarrow\!1$ & .52 & .67 & .62 & .78 & .22 & .74 & .59 \\
$1\!\leftrightarrow\!2$ & .36 & .53 & .49 & .37 & .25 & .49 & .41 \\
$2\!\leftrightarrow\!3$ & .44 & .59 & .57 & .46 & .42 & .54 & .50 \\
$3\!\leftrightarrow\!6$ & .26 & .31 & .34 & .23 & .40 & .24 & .29 \\
\midrule
\multicolumn{8}{l}{\emph{Action Jet trained at $\delta{=}1\sigma$, held-out cosine at probe scale $0.25/0.5/1/2/3/6$}}\\
 & .43 & .58 & \textbf{.82} & .43 & .23 & .11 \\
\bottomrule
\end{tabular}
\caption{Full curvature ladder (DepthSplat, DL3DV, 148 states): per-axis cosine
between per-unit fields at consecutive probe magnitudes, and an Action Jet trained at $\delta{=}1\sigma$ evaluated against every interval. Curvature tracks image flow per unit action, while
$r_y$ (pan), the largest-flow axis, is most curved, and the Action Jet peaks sharply at the displacement it was trained on and collapses off-scale.}
\label{tab:ladder_full}
\end{table}

Table~\ref{tab:controls} lists each control and the
artifact it would expose. The random-field floor ($0.001$) prices dimensionality and the shared autoencoder, where two unrelated fields in this latent space score zero, so a
cosine of $0.80$ is not a high-dimensional artifact. The
seed-paired repeat ($0.573$ on Bridge, $1.000$ on DL3DV) is the ceiling every agreement is read against, and Eq.~\ref{eq:rho_bound} supplies it where a repeat is unaffordable. The per-state $6{\times}6$ mixing fit ($0.12$ against $0.78$ on the DL3DV pair) and the kinematic convention certification ($+1.0000$ against $-0.72$ for the identity map) together close the obvious escape route
from the Bridge negative, namely that the two generators agree but express it in different coordinates.
The copy-frame control is the baseline the validity radius is measured against (Table~\ref{tab:bridge-curve}); $12.94$\,dB is its PSNR on the anchoring band of Table~\ref{tab:bridge_contrast}. And the oracle-of-$K$ ceiling
($0.006$) prices breadth itself, so the selection results of
\S\ref{sec:beyond} report how much of an available gain each verifier captures rather than an unanchored MSE.

\begin{table}[h]
\centering\footnotesize
\begin{tabular}{llr}
\toprule
control & artifact it would expose & measured \\
\midrule
random-field floor & cosine inflated by dimensionality or the shared VAE & $0.001$ \\
seed-paired repeat, same generator & disagreement that is only sampling noise & $0.573$ \\
split-half reliability ceiling $\sqrt{\rho}$ & a stochastic generator penalized for its noise & $0.89$--$0.94$ \\
per-state $6{\times}6$ mixing fit & disagreement recoverable by a linear relabelling & $0.12$ vs.\ $0.78$ \\
kinematic convention certification & generators compared in incompatible action frames & $+1.0000$ vs.\ $-0.72$ \\
rolled-state null & an Action Jet emitting a generic, state-independent field & $0.00$--$0.09$ \\
rolled-action control & anchoring that any two arrays would show & $-0.014$ \\
copy-frame control & a prediction no better than repeating the input & $12.94$\,dB \\
oracle-of-$K$ ceiling & selection gains that breadth alone would deliver & $0.006$ \\
certified null arm & an effect that survives a chart known to be wrong & $-0.011$ \\
reconstruction bound & treating re-rendering methods as competitors & upper bound \\
\bottomrule
\end{tabular}
\caption{Controls, and the artifact each one rules out. Every agreement measurement in the paper is reported against a floor and, where the generator is stochastic, a ceiling.}
\setlength{\tabcolsep}{4pt}
\label{tab:controls}
\end{table}

Table~\ref{tab:students} reports each geometry encoder's
held-out field cosine against its own generator, with a rolled-state null beside it. The nulls sit at $0.00$--$0.09$, so none of these encoders is emitting a generic state-independent field. The DL3DV charts score about $1.5\times$ the Bridge charts, and the reason is the generators rather than
the encoders. On DL3DV both generators are deterministic
(App.~\ref{app:agree_noise}), so their probe targets carry no sampling noise and a perfect encoder would score $1.0$. On Bridge WorldGym is stochastic with measured reliability $\rho_{\mathrm{WG}} = 0.79$--$0.89$, which caps any chart at $\sqrt{0.79} = 0.89$, so its learned chart reaches $0.54$, or about $61\%$ of a ceiling. Table~\ref{tab:students} is measured on a different part of the evaluation split, while App.~\ref{app:agree_bridge} and Table~\ref{tab:bridge-agree} use the 148 agreement states, so their per-axis ranges differ. The same argument cannot be run for IRASim without circularity, since its reliability is bounded from this quantity (Eq.~\ref{eq:rho_bound}).

The DepthSplat chart scores $0.82$ on the full window and $0.73$ on the square one, the same chart and the same
generator, differing only in where it is scored, so the only clean cross-generator row pair here is DepthSplat-square against LagerNVS-square, $0.73$ against $0.77$. This is the same window  the cross-generator protocol of App.~\ref{app:protocol} imposes, and the $0.09$ swing is a measure of what ignoring it would cost. Encoder latency is $8.3$--$15.9$\,ms per state, 
up to $42\%$ of the $37.9$\,ms single-candidate cost of Table~\ref{tab:bridge_nfe}, and the correction head is the other network pass, and everything after it is arithmetic.

\begin{table}[h]
\centering\small
\begin{tabular}{llcccccc|cc}
\toprule
Action Jet & window & $t_x$ & $t_y$ & $t_z$ & $r_x$ & $r_y$ & $r_z$ & mean & null \\
\midrule
DL3DV / DepthSplat & full & .80 & .82 & .79 & .86 & .83 & .84 & \textbf{.82} & .01 \\
DL3DV / LagerNVS & square & .73 & .78 & .73 & .82 & .79 & .78 & \textbf{.77} & .00 \\
DL3DV / DepthSplat & square & .69 & .73 & .69 & .78 & .75 & .74 & \textbf{.73} & .00 \\
Bridge / IRASim & -- & .65 & .68 & .60 & .41 & .44 & .30 & \textbf{.51} & .04 \\
Bridge / WorldGym & -- & .55 & .56 & .62 & .51 & .54 & .47 & \textbf{.54} & .09 \\
\bottomrule
\end{tabular}
\caption{Geometry-encoder held-out field cosine against the Action Jet's own generator, with rolled-state nulls at right. DL3DV Action Jets score about $1.5\times$ Bridge
Action Jets, tracking generator determinism and conditioning tightness. Encoder latency is $8.3$--$15.9$\,ms per state.}
\label{tab:students}
\end{table}

\section{Why the Jacobian has to be measured}
\label{app:illposed}

We tested the natural alternative to a dedicated encoder, namely read-out heads on the frozen generator's features, at several depths and capacities. All of them fail, and the failure is diagnostic rather than an optimization artifact: the map from
generator features to 
the per-block field is ill-posed, in that blocks carrying near-identical generator features require nearly unrelated fields.
Nothing that reads those features pointwise can separate them. Permitting gradient to flow into the generator does reshape its features, at a measurable cost to its
predictive accuracy, and still does not supply the missing structure. A model trained to predict action-conditioned frames is under no obligation to represent
$\partial\,\mathrm{output}/\partial\,\mathrm{action}$ explicitly, and it does not. This is the precise sense in which our method is not output distillation, because there is no representation in the generator to copy, so the Jacobian is obtained by measurement.

Why probe a black box rather than differentiate it. On a
differentiable generator one could read exact Jacobians by automatic differentiation, and as an ablation we verify that the symmetric probe agrees with autodiff where both are available (\S\ref{sec:kubric}), but we do not require it. The
probe treats the generator as a black box, so any model mapping (state, action) to a prediction can serve, including API-only models and multi-step samplers, and the finite difference is a mollified derivative that averages generator noise over a
small but finite $\delta$ rather than differentiating it at a point.

Dispensing with the generator and regressing $z' - z$ on observed action deltas founders on identifiability, where real deltas are correlated and multi-factor, so per-factor fields
entangle exactly where factors co-occur, which is where the geometry is most interesting. The generator answers isolated single-axis queries at arbitrary states, queries the dataset never contains, and that isolation is what makes the recovered
$\widehat{J}_i$ a per-factor derivative rather than a regression coefficient on correlated inputs. %

Parity separation (App.~\ref{app:formal}) is identifiable only if, conditioned on the state, training deltas occur with both signs. Under history conditioning this fails silently, where the context pair reveals the ongoing motion and every training pair
continues it, so the per-state delta distribution is one-sided even when signs balance marginally across clips. Fitted odd and even coefficients then entangle, correct along the observed continuation and arbitrary along the reversal, and query quality becomes direction-asymmetric, which we observed directly as growing artifacts when sliding against a clip's own motion. Plain time reversal does not fix it, because reversing flips context and delta together and leaves counter-momentum queries untrained. 

\section{Metric validity for policy consistency}
\label{app:metric}
Write $\pi$ for a frozen policy and $f_{t+h}$, $\hat f_{t+h}$ for real and predicted futures, and define
$\mathrm{ADE} = \tfrac1H\sum_h \norm{\pi(\hat f_{t+h}) - \pi(f_{t+h})}_2$ under paired sampling keys, so the policy's own sampling noise cancels from the comparison. The obvious alternative, comparing the policy's action on the predicted future against the demonstrated action, may seem equivalent at first glance, but it is not. If we consider what a perfect simulator scores, on the metric we use it scores zero, because predicted and real futures are then
identical, but on the demonstrated-action metric it scores the policy's own cloning error and does not optimize the metric at all. This is evidenced by the fact that our copy-frame control beats the real frames on it ($1.93$ against $2.13$) by exploiting the temporal smoothness of demonstrations. A metric whose optimum is not at the true
simulator cannot rank world models, so we report it as context only.

\begin{table}[h]
\centering\small
\begin{tabular}{l cc cc c}
\toprule
& \multicolumn{2}{c}{DL3DV (LagerNVS chart)} & \multicolumn{2}{c}{DL3DV (DepthSplat chart)} & Bridge (IRASim chart) \\
\cmidrule(lr){2-3}\cmidrule(lr){4-5}\cmidrule(lr){6-6}
axis & $r_1^{*}$ & fill & $r_1^{*}$ & fill & $r_1^{*}$ / fill \\
\midrule
$t_x$ & 0.80 {\scriptsize[0.68,0.92]} & 1.2 & 0.88 {\scriptsize[0.73,1.07]} & 1.4 & 1.89 {\scriptsize[1.65,2.14]} / 0.71 \\
$t_y$ & 0.53 {\scriptsize[0.45,0.64]} & 1.3 & 0.55 {\scriptsize[0.50,0.68]} & 1.3 & 2.10 {\scriptsize[1.83,2.44]} / 0.30 \\
$t_z$ & 1.16 {\scriptsize[1.02,1.34]} & 1.1 & 1.21 {\scriptsize[1.08,1.54]} & 1.4 & 2.13 {\scriptsize[1.79,2.39]} / 0.97 \\
$r_x$ & 0.47 {\scriptsize[0.45,0.50]} & 8.7 & 0.48 {\scriptsize[0.44,0.52]} & 8.5 & 3.33 {\scriptsize[2.80,4.04]} / 0.18 \\
$r_y$ & 0.44 {\scriptsize[0.42,0.46]} & 3.5 & 0.45 {\scriptsize[0.41,0.47]} & 3.7 & 3.02 {\scriptsize[2.44,3.68]} / 0.10 \\
$r_z$ & 0.43 {\scriptsize[0.41,0.45]} & 6.3 & 0.42 {\scriptsize[0.41,0.44]} & 8.1 & 5.63 {\scriptsize[5.00,6.36]} / 0.04 \\
\bottomrule
\end{tabular}
\caption{Self-reported validity radii $r_1^{*}$ (median [IQR]) and stochastic fill floors per unit radius, in normalized chart units where $1.0$ is one probe
displacement. The normalization is what makes the two embodiments comparable.}
\label{tab:radius}
\end{table}

\section{Goal-image verification under a compute budget}
\label{app:budget}

\label{sec:budget}
Planning utility is fidelity times the breadth of futures a planner can afford to weigh, and the cost structure lets us measure the trade. For each state a pool of candidate action chunks is scored by each world model as a goal-image verifier, where we predict every outcome, rank by similarity to the true $t{+}4$ frame, and judge the selected candidate against the demonstrated action, a ground truth no learned proxy can bias, with random-of-$K$ and oracle-of-$K$ bracketing every
curve (Table~\ref{tab:bridge_budget}). 
Breadth pays for competent verifiers, where the oracle and our chart improve with pool size, IRASim holds level, and the oracle-of-$K$ ceiling halves from $K{=}4$ to $16$, which prices breadth itself.
Verification skill dissociates from simulation fidelity, where WorldGym anti-selects, its choices worsening as the pool grows despite mid-pack pixel metrics, while the chart at $150\times$ less cost per candidate matches IRASim at small pools and passes it at
$K{=}32$. Under a fixed budget the ranking inverts, since one second buys IRASim a fifth of a candidate and the chart its entire pool. On a wider pool of 128 candidates the composed
protocol, which stays inside the radius, edges out single extraction at every pool size 
($0.045$ against $0.051$ at $K{=}128$, and this experiment uses its own separate candidate pool and state set, so its values are not comparable with Table~\ref{tab:bridge_budget}), and averaging over noise draws changes
nothing, consistent with the low fill floors on this domain.

\begin{table}[t]\centering\scriptsize
\setlength{\tabcolsep}{3.2pt}
\begin{tabular}{llr ccc c c}
\toprule
method & $g$ & $n$ & PSNR$\uparrow$ & SSIM$\uparrow$ & LPIPS$\downarrow$ & tail LPIPS$\downarrow$ & $\Delta$LPIPS vs.\ ours [95\% CI]\\
\midrule
copy-frame (control) & 1 & 370 & 18.63 & 0.506 & .122 & .220 & $[-0.045,-0.035]$\\
 & 2 & 370 & 16.02 & 0.372 & .205 & .347 & $[-0.008,+0.005]$\\
 & 3 & 370 & 14.83 & 0.313 & .267 & .422 & $[+0.011,+0.025]$\\
 & 4 & 370 & 14.11 & 0.282 & .315 & .471 & $[+0.025,+0.039]$\\
\midrule
DepthSplat & 1 & 370 & 25.64 & 0.862 & .047 & .076 & $[-0.119,-0.110]$\\
 & 2 & 370 & 23.17 & 0.800 & .075 & .125 & $[-0.138,-0.127]$\\
 & 3 & 370 & 21.13 & 0.744 & .104 & .171 & $[-0.151,-0.138]$\\
 & 4 & 370 & 19.55 & 0.695 & .133 & .212 & $[-0.157,-0.142]$\\
\midrule
LagerNVS & 1 & 370 & 30.45 & 0.930 & .039 & .057 & $[-0.128,-0.119]$\\
 & 2 & 370 & 27.38 & 0.879 & .062 & .099 & $[-0.151,-0.139]$\\
 & 3 & 370 & 25.24 & 0.826 & .088 & .142 & $[-0.167,-0.154]$\\
 & 4 & 370 & 23.65 & 0.779 & .115 & .181 & $[-0.175,-0.159]$\\
\midrule
FrameCrafter & 1 & 370 & 17.43 & 0.447 & .187 & .184 & $[+0.029,+0.064]$\\
 & 2 & 370 & 15.36 & 0.343 & .296 & .282 & $[+0.108,+0.162]$\\
 & 3 & 370 & 14.34 & 0.304 & .366 & .375 & $[+0.142,+0.202]$\\
 & 4 & 370 & 13.66 & 0.277 & .413 & .466 & $[+0.157,+0.222]$\\
\midrule
Ours (LagerNVS chart) & 1 & 370 & 20.40 & 0.600 & .162 & .217 & --\\
 & 2 & 370 & 18.67 & 0.507 & .207 & .283 & --\\
 & 3 & 370 & 17.50 & 0.438 & .249 & .338 & --\\
 & 4 & 370 & 16.76 & 0.390 & .283 & .377 & --\\
\midrule
Ours (DepthSplat chart) & 1 & 370 & 20.48 & 0.597 & .164 & .217 & $[+0.001,+0.003]$\\
 & 2 & 370 & 18.69 & 0.503 & .211 & .286 & $[+0.002,+0.005]$\\
 & 3 & 370 & 17.53 & 0.435 & .253 & .342 & $[+0.002,+0.005]$\\
 & 4 & 370 & 16.78 & 0.387 & .287 & .381 & $[+0.003,+0.006]$\\
\bottomrule
\end{tabular}
\caption{All methods, causal tier (two trailing context frames), common 256-pixel window. The copy control and the difficulty strata use exactly our states. Tail LPIPS averages the hardest third of states, ranked by how much the copy control misses. $\Delta$LPIPS is paired (row minus ours) on shared states; negative means better than ours. An interpolation tier (six surrounding context views) is not reported: its context includes the target frame at $g{\ge}2$.}
\label{tab:nvs_all}
\end{table}

\begin{table}[t]\centering\small
\setlength{\tabcolsep}{3.2pt}
\begin{tabular}{l cccc cccc}
\toprule
& \multicolumn{4}{c}{PSNR$\uparrow$} & \multicolumn{4}{c}{LPIPS$\downarrow$}\\
\cmidrule(lr){2-5}\cmidrule(lr){6-9}
gap $g$ & 1 & 2 & 3 & 4 & 1 & 2 & 3 & 4\\
\midrule
\multicolumn{9}{l}{\emph{LagerNVS generator}}\\
single extraction, FT decoder & 21.2 & 19.7 & 18.5 & 17.6 & .139 & .172 & .209 & .243\\
composed (AR), FT decoder & 21.2 & 18.6 & 16.8 & 15.5 & .139 & .217 & .284 & .344\\
composed (AR), frozen decoder & 21.0 & 18.0 & 16.1 & 14.8 & .158 & .273 & .388 & .478\\
\midrule
\multicolumn{9}{l}{\emph{DepthSplat generator}}\\
single extraction, FT decoder & 21.4 & 19.7 & 18.5 & 17.6 & .142 & .177 & .214 & .249\\
composed (AR), FT decoder & 21.4 & 18.9 & 17.1 & 15.9 & .142 & .222 & .292 & .352\\
composed (AR), frozen decoder & 21.1 & 18.2 & 16.3 & 15.0 & .158 & .281 & .410 & .505\\
\midrule
full window, frozen decoder (DepthSplat generator) & 17.7 & 16.9 & 16.3 & 15.6 & .213 & .285 & .350 & .407\\
\bottomrule
\end{tabular}
\caption{Ablations of our chart on the $n{=}100$ states per gap on all four gaps. Composing four single-step queries autoregressively is worse than one extrapolated query at $g{\ge}2$ (paired $\Delta$LPIPS $+0.045$ to $+0.103$, every CI excluding zero): re-encoding each decoded frame compounds error faster than the calibrated jet loses accuracy in extrapolation. The fine-tuned decoder's gain grows under composition (LPIPS $+0.016$ at $g{=}1$ to $+0.153$ at $g{=}4$, frozen minus fine-tuned), because every step re-encodes the previous decode. At $g{=}1$ both extraction modes coincide exactly, a pipeline-identity check. The full-window row is not decoder-matched and is shown for feasibility only.}
\label{tab:nvs_abl}
\end{table}

\begin{table}[h]
\centering
\scriptsize
\begin{tabular}{llcccccccc}
\toprule
 & decoder & \multicolumn{2}{c}{gap 1} & \multicolumn{2}{c}{gap 2}
 & \multicolumn{2}{c}{gap 3} & \multicolumn{2}{c}{gap 4} \\
\midrule
free-form        & frozen & 33.54 & .0325 & 30.32 & .0638 & 27.53 & .0942 & 26.24 & .1249 \\
free-form        & FT     & 34.45 & .0153 & 31.19 & .0267 & 28.28 & .0391 & 27.26 & .0511 \\
free-form+pixFT  & frozen & 32.45 & .0285 & 28.73 & .0558 & 25.96 & .0884 & 24.54 & .1275 \\
free-form+pixFT  & FT     & 33.49 & .0165 & 30.17 & .0274 & 27.72 & .0381 & 26.77 & .0484 \\
2-jet+pixFT      & frozen & 28.69 & .0332 & 26.09 & .0680 & 23.65 & .1033 & 22.76 & .1373 \\
2-jet+pixFT      & FT     & 30.46 & .0213 & 27.93 & .0350 & 25.86 & .0468 & 25.33 & .0575 \\
3-jet            & frozen & 28.98 & .0319 & 26.31 & .0643 & 24.07 & .0952 & 23.13 & .1272 \\
3-jet            & FT     & 31.80 & .0183 & 29.05 & .0305 & 26.92 & .0403 & 26.27 & .0501 \\
\bottomrule
\end{tabular}
\caption{Kubric full ablation ladder. Correction-head ladder $\times$ decoder (PSNR / LPIPS per gap). ``+pixFT'' marks training with optional adversarial loss. The shared decoder was fine-tuned on \{3-jet, free-form+pixFT\} latents;
the out-of-mix plain free-form model rendering best through it
evidences that the shared renderer plays no favorites.}
\label{tab:kubric_ablation}
\end{table}

\end{document}